\documentclass[a4paper]{cas-dc}

\usepackage[numbers]{natbib}

\newcommand{\etal}{\textit{et al. }}
\newcommand{\tabincell}[2]{\begin{tabular}{@{}#1@{}}#2\end{tabular}}
\usepackage{color}
\usepackage{setspace}

\def\tsc#1{\csdef{#1}{\textsc{\lowercase{#1}}\xspace}}
\tsc{WGM}
\tsc{QE}
\begin{document}
\let\WriteBookmarks\relax
\def\floatpagepagefraction{1}
\def\textpagefraction{.001}

% Short title
\shorttitle{A Survey on Deep Learning-based Single Image Crowd Counting: Network Design, Loss Function and Supervisory Signal}    

% Short author
\shortauthors{Haoyue Bai et al.}  

% Main title of the paper
\title [mode = title]{A Survey on Deep Learning-based Single Image Crowd Counting: Network Design, Loss Function and Supervisory Signal}  

% Title footnote mark

\author[1]{Haoyue Bai}[type=editor,
       orcid=0000-0001-8139-0431
       ]

\affiliation[1]{organization={The Hong Kong University of Science and Technology},
            city={Hong Kong}
            }

\author[2]{Jiageng Mao}

\affiliation[2]{organization={The Chinese University of Hong Kong},
            city={Hong Kong}
            }
            
\author[1]{S.-H. Gary Chan}

% Here goes the abstract
\begin{abstract}
\noindent Single image crowd counting is a challenging computer vision problem with wide applications in public safety, city planning, traffic management, etc. With the recent development of deep learning techniques, crowd counting has aroused much attention and achieved great success in recent years.
This survey is to provide a comprehensive summary of recent advances on deep learning-based crowd counting techniques via density map estimation by systematically reviewing and summarizing more than 200 works in the area since 2015.
%In this work, we have systematically reviewed and summarized more than 200 crowd counting works using deep learning methods since 2015.
Our goals are to provide an up-to-date review of recent approaches, and educate new researchers in this field the design principles and trade-offs.
After presenting publicly available datasets and evaluation metrics, 
we review the recent advances with detailed comparisons on
three major design modules for crowd counting:
deep neural network designs, loss functions, and supervisory signals.
We study and compare the approaches using the public datasets and evaluation metrics.
We conclude the survey with some future directions.
\end{abstract}


\maketitle

% Main text

%-----------------------------Introduction-------------------------------

\section{Introduction}
Single image crowd counting is to estimate the number of objects
(people, cars, cells, etc.) in an image of an unconstrained scene, i.e.,
an image without any restriction on the scene.
Crowd counting has attracted much attention in recent years due to its important applications in public safety, traffic management, consumer behavior, cell counting, etc.~\cite{onoro2016towards, lempitsky2010learning, chan2008privacy}.
In this survey, we mainly focus on people as the crowd, though the techniques discussed may be extended to other domains.

Due to the importance of crowd counting, extensive research have been done in the area, especially with the use of deep learning, which has demonstrated superior performances on various applications, such as computer vision~\cite{he2016deep, mao2021pyramid, mao2021one}, image classification~\cite{krizhevsky2012imagenet},
%natural language processing~\cite{devlin2018bert}, 
and multi-dimensional time series~\cite{aydin2019deep}. %Compared with machine learning models, 
Deep learning achieves success for single image crowd counting with large-scale publicly available benchmarks~\cite{idrees2018composition, wang2020nwpu} in recent years. This may be due to its data-driven properties~\cite{zhang2016single, li2018csrnet} and capability of self-learning from raw data~\cite{liu2018leveraging, sam2019almost} for deep learning-based methods. In this work, we mainly discuss recent advanced deep learning-based single image crowd counting approaches due to its superiority in comparison to machine learning models.

Early approaches to count people are based on detection-based computer vision techniques, which are to detect individual objects, heads, or body parts and then count the total number in the image~\cite{rabaud2006counting, lin2010shape, li2008estimating}. However, its accuracy deteriorates quickly for crowded scenes where objects have severe occlusions. To overcome it, the regression-based approach has been recently proposed, which directly estimates the count by relating it with the image.
While achieving higher accuracy than the detection-based approach for crowded scenes,
it
lacks adequate spatial information of the people and is less interpretable~\cite{chan2012counting, wang2015deep, chan2009bayesian}, hindering its extension to localization study.

\begin{figure*}[t]
	\centering
	\includegraphics[width=1.0\textwidth]{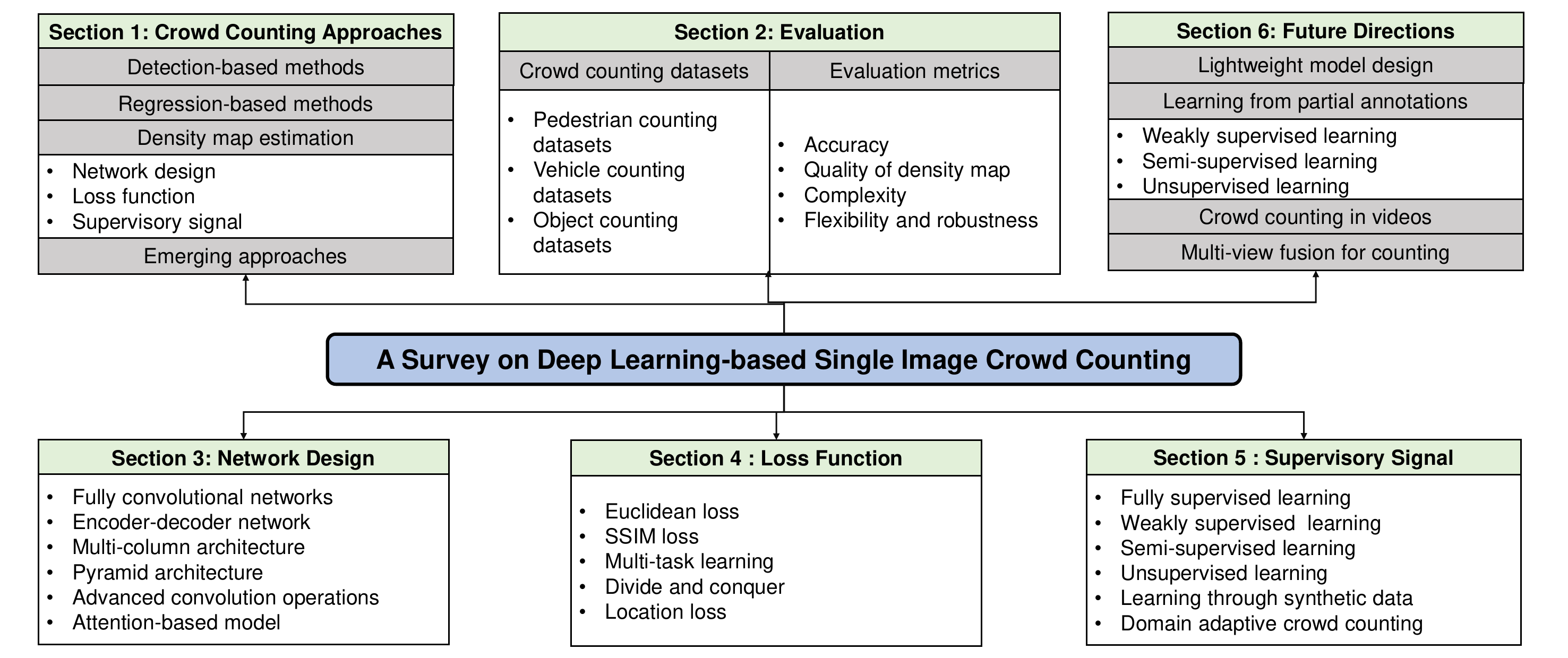}
	\caption{ The structure of this survey. First, we overview the four main categories of deep learning-based crowd counting methods. Second, we present publicly available counting datasets and evaluation metrics. Then, we review recent advances on crowd counting schemes, which is mainly pertain to deep neural network design, loss function and supervisory signal. We conclude the survey with future directions.}
	\label{fig:structure}
\end{figure*}

Most recently, crowd counting via density map estimation has emerged as a promising approach with encouraging results, where the input image is processed to a crowd density map, 
which is simply integrated to obtain the number of people in a pixel of the image
~\cite{lempitsky2010learning, pham2015count, bai2019crowd, cao2018scale, zhang2016single, liu2019adcrowdnet, li2018csrnet, huberman2022single}.
Such approaches achieve high accuracy for crowded scenes and preserve spatial information of people distribution.
Besides, there are some emerging approaches such as S-DCNet~\cite{wang2019learning} which classifies the features into a predefined count range for crowd estimation.

We summarize by comparing the four major crowd counting approaches in Table~\ref{tab:4category}.  All of them require image annotation through labeling in the training step.  For detection-based approach, each object has to be fully identified and outlined, which incurs the highest labeling cost.  On the other hand, regression-based approach does not need to annotate individual objects but the total object count, and hence its annotation cost is the lowest.  Density %map 
estimation has an intermediate labeling cost between the two because only the heads of the people need to be indicated.

\begin{figure*}[t]
	\centering
	\includegraphics[width=1.0\textwidth]{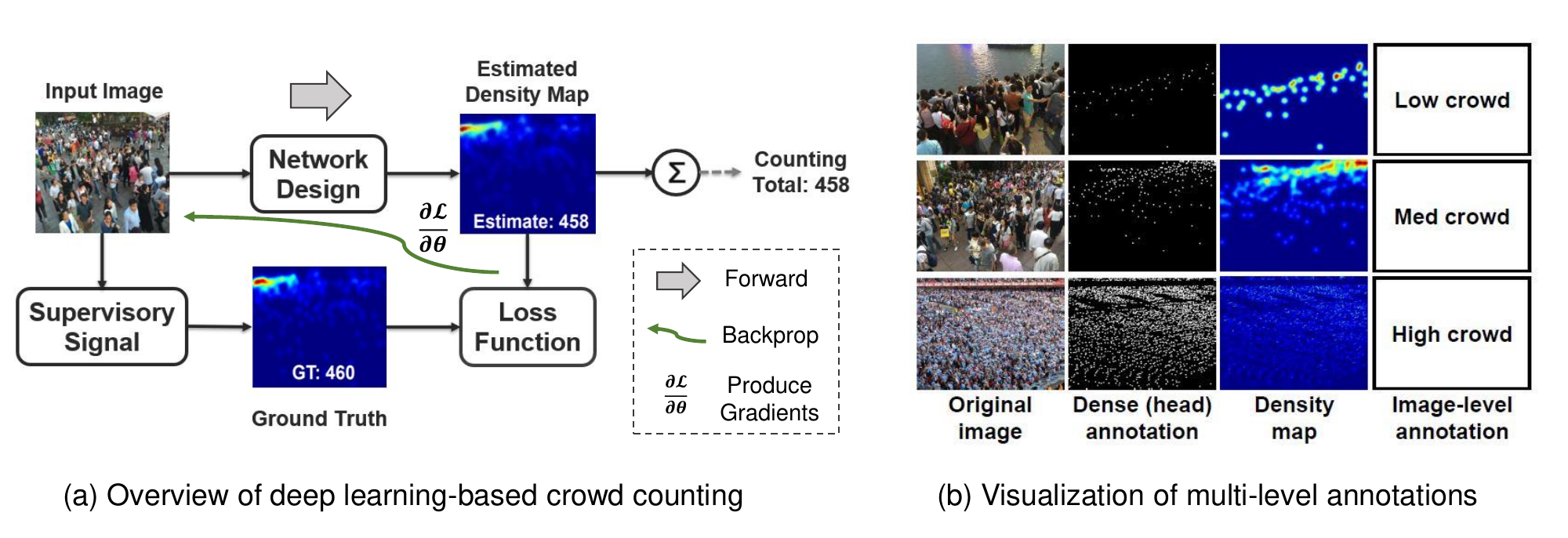}
	%\caption{Overview of CNN-based single image crowd counting methods via density map estimation.}
	\caption{ Overview of deep learning-based single image crowd counting methods via density map estimation. Figure (a) shows the major components for deep learning-based crowd counting via density estimation. Figure (b) presents visualization of original image, labor-intensive dense annotation, ground truth density maps, and image-level weak annotation. The annotation paradigms are from~\cite{sindagi2019ha}.}
	\label{fig:flow}
\end{figure*}

We focus in this survey on crowd counting via density map estimation.
With the development of deep learning approaches in the field of computer vision, counting accuracy has been greatly improved with the use of 
deep learning-based
models as compared with approaches based on handcrafted features.
We overview in Fig.~\ref{fig:flow} (a) the major design components for
CNN-based crowd counting via density map estimation.
An input image of a crowd scene is fed into a deep neural network which 
estimates the density map of the image
(the upper branch).  Here the critical issue is the {\em network design} so that the sum of the density value in all the pixels closely matches with the crowd count in the input image. For training (the lower branch),
an image is first annotated with {\em supervisory signal}, which 
may range from fully to pseudo labeled, to generate the ground truth %density map 
(given by the number of people per pixel in the image).
The ground truth is used to adjust the node parameters of the deep neural network through minimizing a {\em loss function} between the network-generated density map and the ground truth.

We present recent advances on deep learning-based crowd counting.
%CNN-based crowd counting.  
Our goals are to educate the new researchers state-of-the-arts and equip them with insights, tools, and principles to design novel networks.
We survey and compare
the available datasets, performance metrics,
network design, loss function, and supervisory signal. Our survey is timely and unique. 

We discuss %previous 
related work as follows.
Teixeira~\etal~\cite{teixeira2010survey} is an early survey on human sensing. However, it has not focused on crowd scene analysis.
Li~\etal~\cite{li2014crowded} reviews crowd scene analysis in terms of crowd behavior, activity analysis, and anomaly detection, with crowd counting playing a small role.
Ahuja~\etal~\cite{ahuja2019survey} covers different crowd estimation methods.
Though Zitouni~\etal~\cite{zitouni2016advances} evaluate different crowd analysis methods,
is not mainly for CNN-based approach via density map estimation, which has become the mainstream for crowd counting in recent years.
Chrysler~\etal~\cite{chrysler2021literature} discusses the methods to tackle the challenges of the lack of training data, perspective distortion faced by the crowd counting system.
%Sindagi et al
The work~\cite{sindagi2018survey} surveys on CNN-based %crowd counting 
approach for a single image, but
it only roughly discussed the recent advances on CNN-based methods.
It has not discussed the advanced convolutional operations and attention-based model, loss function, and supervisory signal, and only up to the year 2017.
%Gao~\etal~\cite{gao2020cnn} focus on the modern CNN-based density estimation and crowd counting approaches. However, it has not covered the hot recent advances on network design (e.g., vision transformers~\cite{dosovitskiy2020image, liu2021swin}), loss function, challenges on lack of training data as we discussed here.

\begin{table*}[h]
    \centering
    \caption{
    Summary of crowd counting approaches on four major categories: detection-based, regression-based, density map estimation, and emerging approaches.}
    \begin{tabular}{|c|c|c|c|c|c|c|c|}
        \hline
        \textbf{Category} &\textbf{Principles}
        &\textbf{\tabincell{c}{Crowd\\Counting\\Accuracy}}
        &\textbf{\tabincell{c}{Location\\Accuracy}}
        &\textbf{\tabincell{c}{Annotation\\Complexity}}
        &\textbf{Limitations} &\textbf{ Examples}
        \\
        \hline
        %\hline
        \tabincell{c}{Detection\\based}
        &\tabincell{c}{Detect\\then count;\\early approach}
        &Low &High &\tabincell{c}{High\\(object framing)}
        &\tabincell{c}{Low accuracy\\for highly\\crowded scenes} 
        & \tabincell{c}{~\cite{rabaud2006counting},~\cite{lin2010shape},\\\cite{li2008estimating}}
        \\
        \hline
        \tabincell{c}{Regression\\based}
        &\tabincell{c}{Directly learn\\to regress\\the count}
        &Medium &N/A &\tabincell{c}{Low\\(image-level)}
        &\tabincell{c}{Less interpretable;\\lacks location\\information} 
        &{
        \tabincell{c}{~\cite{chan2012counting},~\cite{wang2015deep},\\\cite{chan2009bayesian}}}
        \\
        \hline
        \tabincell{c}{\\Density map\\estimation}
        &\tabincell{c}{Compute number\\of people\\per pixel}
        &High &Medium &\tabincell{c}{Medium\\(head indication)}
        &\tabincell{c}{Low accuracy\\in low\\crowd scenes}
        &{
        \tabincell{c}{~\cite{lempitsky2010learning},~\cite{pham2015count},\\\cite{cao2018scale},~\cite{zhang2016single},\\\cite{liu2019adcrowdnet},~\cite{li2018csrnet}}}
        \\
        \hline
        \tabincell{c}{{ Emerging}\\{approaches}}
        &{
        \tabincell{c}{Classify the\\features into\\a predefined\\count range}}
        &{ High} &{ Low} 
        &{
        \tabincell{c}{Medium\\(head indication)}}
        &{
        \tabincell{c}{Not flexible\\to wide\\count range}}
        &{
        \tabincell{c}{~\cite{xiong2019open}}}
        \\
        \hline
    \end{tabular}
    \label{tab:4category}
\end{table*}

\begin{table*}[ht]
    \centering
    %\doublespacing
    \footnotesize
    \caption{A comprehensive analysis of other counting related survey papers. Compared with previous related works, our work is of current interest and value, because it is timely, more comprehensive and provide an in-depth analysis of the representative approaches in this active area.
    }
    %\doublespacing
    \begin{tabular}{|l|c|c|c|c|}
        \hline
        \textbf{ Paper} &\textbf{ Year}
        &\textbf{ Venue}
        &\textbf{Comparison of Other Crowd Counting Surveys}
        \\
        \hline
        %\hline
        {
        \tabincell{l}{Approaches on Crowd Counting and\\Density Estimation: A Review~\cite{li2021approaches}}
        }
        &{ 2021}
        &{ PAA}
        &
        {
        \tabincell{c}{This paper focus on elaborating deep\\learning-based counting methods, which\\is board-based and mainly focus on the network\\design considerations without discussing loss\\functions and supervisory signals.
        }}
        \\
        \hline
        {
        \tabincell{l}{A Literature Review of\\Crowd-counting System on\\Convolutional Neural Network~\cite{chrysler2021literature}}}
        &{ 2021}
        &{IOPCS}
        &
        {
        \tabincell{c}{This survey discusses the challenges faced by crowd\\ counting systems and focuses on developing a more robust\\crowd counting methodology. However, this survey is\\a short paper. The network design discussion misses\\some important recent approaches such as\\DM-Count, SASNet. It also lacks unsupervised\\learning counting approaches.
        }}
        \\
        \hline  
        {
        \tabincell{l}{A Survey of Recent Advances\\in Crowd Density Estimation\\using Image Processing~\cite{ahuja2019survey}}
        }
        &{ 2019}
        &{ICCES}
        &
        {
        \tabincell{c}{This is a short paper, which mainly discusses\\the traditional approaches with\\hand-crafted features. Deep learning-based\\approaches only play a small part.
        }}
        \\
        \hline   
        {
        \tabincell{l}{A Survey of Techniques\\for Automatically Sensing\\the Behavior of a Crowd~\cite{draghici2018survey}}}
        &{ 2018}
        &{ACMCS}
        &
        {
        \tabincell{c}{This paper surveys practical solutions for sensing\\ pedestrian behavior, and also combining privacy, transparency,\\ scalability, and ease of deployment. However, this paper is for\\ traditional methods with hand-crafted features.}}
        \\
        \hline   
        {
        \tabincell{l}{A Survey of Recent Advances\\in CNN-based Single Image\\Crowd Counting and\\Density Estimation~\cite{sindagi2018survey}}}
        &{ 2018}
        &{PRL}
        &
        {
        \tabincell{c}{This paper surveys CNN-based crowd counting approaches\\ for a single image, but it only roughly discussed the recent\\advances on CNN-based methods. It has not discussed the\\advanced convolutional operations and attention-based\\model, loss function and supervisory signal, and\\only up to the year 2017.
        }}
        \\
        \hline
        {
        \tabincell{l}{Crowded Scene Analysis:\\A Survey~\cite{li2014crowded}}
        }
        &{2014}
        &{T-CSVT}
        &
        {
        \tabincell{c}{This paper reviews crowd scene analysis in terms of\\ crowd behavior, activity analysis, and anomaly detection,\\with crowd counting playing a small role.} 
        }
        \\
        \hline
        {
        \tabincell{l}{Advances and Trends in Visual\\Crowd Analysis: A Systematic\\Survey and Evaluation of Crowd\\Modelling Techniques~\cite{zitouni2016advances}}
        }
        &{2016}
        &{ NC}
        &
        {
        \tabincell{c}{This paper evaluates different crowd analysis\\methods,  is not mainly for CNN-based approach\\via density map estimation, which  has become the\\mainstream for crowd counting in recent years.}}
        \\
        \hline
        {
        \tabincell{l}{A Survey of Human-Sensing: Methods\\for Detecting, Presence, Count,\\Location, Track, and Identity~\cite{teixeira2010survey}}
        }
        &{2010}
        &{CS}
        &
        {
        \tabincell{c}{An early survey on human sensing. However, it has not\\ focused on crowd scene analysis but on the study of presence,\\count, location, track, and identification.} 
        }
        \\
        \hline
    \end{tabular}
    \label{tab:rela}
\end{table*}

In contrast with previous papers, our work comprehensively summarizes more than two hundred deep learning-based
%CNN-based 
crowd counting algorithms in the recent five years.
%Figure~\ref{fig:time} presents the existing crowd counting methods with a timeline axis, which summarize the development of this field.
Our work is of current interest and value, because it is more comprehensive, summarizing the more recent, popular, and critical design components of this active field
and provide an in-depth illustration of
the representative schemes in the area. 
Through this survey, we expect to offer an up-to-date summary of recent advances in this field and educate new researchers on the design principles and trade-offs.

Figure~\ref{fig:structure} shows the main design components for crowd counting we will discuss in this paper.
For network design, we describe the basic principles of major techniques such as fully convolutional network, encoder-decoder architecture, multi-column, and pyramid network, etc.
For loss function, we discuss the widely used Euclidean loss and the recently advanced schemes such as SSIM loss, and multi-task learning. For supervisory signal, we introduce different ground truth generation methods for fully supervised setting and compare it with
weakly supervised and semi-supervised learning, %unsupervised 
and self-supervised learning, and automatic labeling through synthetic data.
Typical representative schemes are summarized and compared in each section.

The rest of the paper is organized as follows. In Section~\ref{sec:metric}, we summarize the publicly available crowd counting datasets, evaluation metrics, and design considerations. We present in Section~\ref{sec:net} the details of deep neural network design. Section~\ref{sec:loss} discusses the loss functions, and Section~\ref{sec:data} reviews supervisory signal to train crowd counting network. We conclude with future directions in Section~\ref{sec:conc}.

%----------------------------Data----------------------------------
%\input{chap/data}
\section{Datasets and Performance Evaluation}\label{sec:metric}

In this section, we first summarize the most widely used crowd counting datasets in Section~\ref{met:data}. Then we discuss the design considerations and performance metrics to study crowd counting in Section~\ref{met:eval}.

\subsection{Datasets}\label{met:data}
Public datasets are used as benchmarks to evaluate crowd counting models.
In choosing a dataset, the following metrics are often considered:
\begin{itemize}
\item \textit{Image resolution:} Datasets with high resolutions usually show better visual quality. Furthermore, due to their higher pixel density, they often achieve higher count accuracy.
    %\item \textit{Number of images in the dataset:} 
    \item \textit{Number of images:} 
    Datasets with a large number of images often cover more diverse scenes, backgrounds, view angles, and lighting conditions. Large and diverse datasets are beneficial to optimize deep learning-based models and mitigate over-fitting problems. 
    \item \textit{Object count:} The number of objects in a dataset is an important consideration for crowd analysis. The minimum, maximum, and average counts shed light on %the 
    crowd density in the dataset. Datasets with a large crowd density level coverage
    and the number of objects is usually more challenging for crowd counting.
\end{itemize}

We identify some common datasets used in the research community including pedestrian counting and object datasets, and
extract and present some typical images from the datasets
in Fig.~\ref{fig:dataset}. There are also some other works focus on counting from remote scenes~\cite{zhu2021graph, du2020visdrone, zhao2020flow, gao2020counting, meng2021count} and indoor crowd counting~\cite{ling2019indoor}.
We also 
compare these datasets in Table~\ref{table:dataset}. These datasets are elaborated below:

\begin{figure*}[t]
	\centering
	\includegraphics[width=1\textwidth]{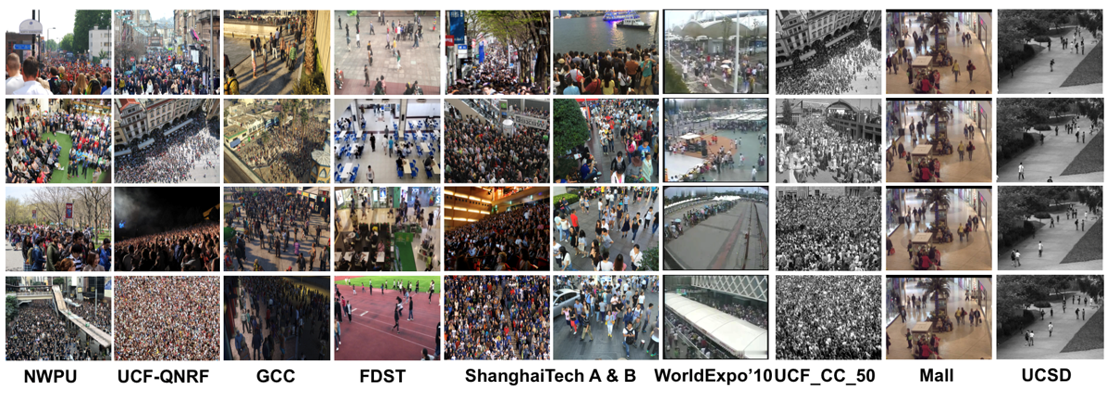}
	\caption{Some typical crowd scene images of publicly available datasets. Different columns represents different crowd counting datasets and we visualize four typical images for each dataset. The NWPU~\cite{wang2020nwpu}, UCF-QNRF~\cite{idrees2018composition}, ShanghaiTech A \& B~\cite{zhang2016single},  WorldExpo'10~\cite{zhang2015cross}, and UCF\_CC\_50~\cite{idrees2013multi} are image-based datasets. The FDST~\cite{fang2019locality}, Mall~\cite{chen2012feature}, and UCSD~\cite{chan2008privacy} are video-based datasets. The GCC~\cite{sam2019almost} is a diverse synthetic crowd dataset. }
	\label{fig:dataset}
\end{figure*}

\begin{itemize}

\item \textit{RGBT-CC} contains RGB-thermal data captured in different scenarios in urban scenes with various densities, e.g., malls, streets, playgrounds, train stations, etc. $1,013$ pairs are in light and $1,017$ pairs are %captured 
in darkness. RGBT-CC is randomly divided into three sets: 1030 pairs for training, 200 pairs for validation, 800 pairs for testing.

\item \textit{NWPU-Crowd}~\cite{wang2020nwpu}
consists of $5,109$ images and $2,133,375$ annotated instances with points and boxes. Compared with other real-world crowd counting datasets, the NWPU-Crowd dataset has the largest density range of the annotated objects from 0 to $20,033$ per image. The average resolution of this dataset is $2191 \times 3209$, which is generally larger than other widely used 2D single image crowd counting datasets.

\item \textit{JHU-Crowd}~\cite{sindagi2019pushing} is collected under diverse scenarios, environmental, and weather conditions include images with weather-based degradations and illumination variations. This dataset contains a rich set of labels: blur-level, occlusion-level, size-level, and other image-level annotations. 

\item \textit{CrowdSurveillance}~\cite{yan2019perspective} is a large scale crowd counting dataset with high-resolution images captured under challenging scenarios. The dataset is built by both online crawling and real-world surveillance video which covers more challenging scenarios with complicated backgrounds and varying crowd counts.

\item \textit{DroneCrowd}~\cite{wen2021detection} is formed by 112 video clips with 33,600 high resolution frames with large variations in scale, viewpoint, and background clutters, which captured under 70 different scenarios across 4 cities. The video clips are recorded at 25 frames per seconds with $1920 \times 1080$ resolution.
This dataset also provides 20,800 people trajectories with head annotations and several video-level attributes in sequences, i.e., illumination, altitude, and density. DroneCrowd is divided into training and testing sets, with 82 and 30 video sequences respectively.

\item \textit{UCF-QNRF}~\cite{idrees2018composition}
contains 1,535 challenging images and a total of 1,251,642 annotations. The minimum and the maximum number of objects within an image are 49 and 12,865. The training and testing sets are selected by sorting the images according to the counts and picking every 5th image as the test set (1201 images for training and 334 images for testing).
Besides, this large-scale dataset covers different locations, viewpoints, perspective effects, and different times of the day.

\item \textit{GCC}~\cite{sam2019almost}~\cite{wang2020pixel}
is a large-scale diverse synthetic crowd dataset, which was generated based on a computer game, Grand Theft Auto V. GTA V Crowd Counting (GCC) dataset consists of $15,212$ images, with a resolution of $1080 \times 1920$, containing more than $7,625,843$ people annotation. GCC is more diverse than other real-world datasets. It captures 400 different crowd scenes in the GTA C game, which includes multiple types of locations.

\item \textit{Fudan-ShanghaiTech}~\cite{fang2019locality} 
contains 100 videos captured from 13 different scenes. FDST includes 150,000 frames and 394,081 annotated heads, which is larger than previous video crowd counting datasets in terms of frames. The training set of the FDST dataset consists of 60 videos, 9000 frames, and the testing set contains the remaining 40 videos, 6000 frames.
The number of frames per second (FPS) for FDST is 30.

\item \textit{ShanghaiTech A $\&$ B}~\cite{zhang2016single}
consists of two parts: Part A and Part B, which contains 482 images (300 images for training, 182 images for testing), and 716 images (400 images for training, 316 images for testing), respectively.
Part A includes high-density crowds that are collected from the Internet. Part B is captured from the busy streets of urban areas in Shanghai, which are less crowded than the scenes from Part A.

\item \textit{WorldExpo'10}~\cite{zhang2015cross}
focus on cross-scene counting. It consists of 1132 video sequences captured by 108 surveillance cameras during the Shanghai 2010 WorldExpo. WorldExpo'10 dataset is randomly selected from the video sequences, which has 3,980 frames with 199,923 head annotations. The training set of WorldExpo'10 contains 3,380 frames from 103 scenes, and the remaining 600 frames are sampled from five other different scenes with each scene being 120 frames for testing.

\item \textit{UCF\_CC\_50}~\cite{idrees2013multi}
has 50 black and white crowd images and 63974 annotations, with the object counts ranging from 94 to 4543 and an average of 1280. The original average resolution of the dataset is $2101 \times 2888$. This challenging dataset is crawled from the Internet. For experiments, UCF\_CC\_50 were divided into 5 subsets and performed five-fold cross-validation. The maximum resolution was reduced to 1024 for efficient computation.

\begin{table*}[t]
	\centering
	%\doublespacing
	\caption{ An overview of datasets statistics for crowd counting. \textbf{Image Number} is the number of images;  \textbf{Total} is total number of labeled objects; \textbf{Min Count} is the minimal crowd count; \textbf{Max Count} is the maximum crowd count; \textbf{Ave Count} is the average crowd count.
    }
    %\doublespacing
	\label{table:dataset}
	\begin{tabular}{|c|l|c|c|c|c|c|c|c|}
		\hline
		\textbf{Category}
		&\textbf{Dataset}       
		&\textbf{Year}   
		&\textbf{\tabincell{c}{Average\\Resolution}} &\textbf{\tabincell{c}{Image\\Number}} &\textbf{Total} 
		&\textbf{\tabincell{c}{Min\\Count}}  &\textbf{\tabincell{c}{Max\\Count}}
		&\textbf{\tabincell{c}{Avg\\Count}}\\
		\hline
		{
		\multirow{15}{*}{\tabincell{c}{Pedestrian\\Counting}}} &{RGBT~\cite{liu2021cross}} &{2021} &{640$\times$480} &{2,030} &{138,389} &{-} &{-} &{68} \\ 
		~ &NWPU-Crowd~\cite{wang2020nwpu}
		&2020 &2191$\times$3209 &5,109 &2,133,375 &0  &20,033 &418\\
		~ &{JHU-Crowd~\cite{sindagi2019pushing}} &{2019} &{1450$\times$900} &{4250} &{1,114,785} &{0}&{7286} &{262} \\
		~ &{\tabincell{c}{Crowd Surveillance~\cite{yan2019perspective}}} &{2019} &{1342$\times$840} &{13,945} &{386,513} &{-} &{-} &{35} \\
		~ &{DroneCrowd~\cite{wen2021detection}}&{2019} &{1920$\times$1080} &{33,600} &{4,864,280} &{25} &{455} &{145} \\
		~ &UCF-QNRF~\cite{idrees2018composition}
		&2019 &2013$\times$2902 &1,535 &1,251,642 &49  &12,865&815 \\
		~ &GCC~\cite{sam2019almost} &2019 &1080$\times$1920  &15,212 &7,625,843 &0  &3,995 &501\\
		~ &\tabincell{c}{Fudan-ST~\cite{fang2019locality}}
		&2019 &1080$\times$1920 &15,000 &394,081 &9  &57 &27\\
		%SmartCity   &2018 &1080$\times$1920 &50                   %&7   
		%&369          \\
		~ &\tabincell{c}{ST Part A~\cite{zhang2016single}}
		&2016 &589$\times$868 &482 &241,677 &33  &3,139 &501\\
		~ &\tabincell{c}{ST Part B~\cite{zhang2016single}}
		&2016 &768$\times$1024 &716 &88,488 &9  &578 &124\\
		~ &WorldExpo'10~\cite{zhang2015cross}
		&2015 &576 $\times$ 720 &3,980 &199,923 &1  &253 &50\\
		~ &UCF\_CC\_50~\cite{idrees2013multi}
		&2013 &2101$\times$2888 &50 &63,974 &94  &4,543 &1,280\\
		~ &Mall~\cite{chen2012feature}
		&2012 &240$\times$320 &2,000 &62,325 &13  &53 &31\\
		~ &UCSD~\cite{chan2008privacy}
		&2008 &158$\times$238 &2,000 &49,885 &11  &46 &25\\
		\hline
		{\multirow{3}{*}{\tabincell{c}{Object\\Counting}}}
		~ &{VisDrone Vehicle~\cite{zhu2018visdrone}}& {2019} & {991$\times$1511} &{5303} &{198,984} &{10} &{349} &{38}\\
		~ &{Penguin~\cite{arteta2016counting}} &{2016} & {700$\times$700}& {8200} &{72160} &{-} &{5} &{8.8} \\	
		~ &{TRANCOS~\cite{guerrero2015extremely}} &{2015} & {640$\times$480} &{1244} &{46796} &{-} &{-} &{38} \\
		\hline
	\end{tabular}
\end{table*}

\item \textit{Mall}~\cite{chen2012feature}
was captured by a public surveillance camera in a shopping mall, which contains more challenging lighting conditions and more severe perspective distortion than the UCSD dataset~\cite{chan2008privacy}. The Mall dataset consists of 2000 video frames with fixed resolution ($320 \times 240$) and 62,325 total pedestrian instances. The first 800 frames were used for training and the remaining 1200 frames for testing.

\item \textit{UCSD}~\cite{chan2008privacy}
consists of an hour of video with 2000 annotated frames and in a total of 49,885 pedestrian instances, which was captured from a pedestrian walkway of the UCSD campus by a stationary camera. The original video was recorded at 30fps with a frame size of $480 \times 740$ and later downsampled to 10fps with dimension $158 \times 238$. The 601-1400 frames were used for training and the remaining 1200 frames for testing. The ROI of the walkway and the traveling direction are also provided.

\item \textit{VisDrone Vehicle}~\cite{bai2019crowd} is modified from the original VisDrone2019 detection dataset~\cite{zhu2018visdrone} with bounding boxes of targets to crowd counting annotations. The new vehicle annotation location is the center point of the original bounding box. This dataset consists of 3953 training samples, 364 validation samples, and 986 test samples.

\item \textit{Penguin}~\cite{arteta2016counting} is a large and challenging %image 
dataset of penguins in the wild with high-degree of object occlusion and scale variation. The collected images are compounded by many factors, e.g., adversarial weather conditions, variability of vantage points of the cameras, extreme crowding, and inter-occlusion between penguins. The Penguin dataset is divided into two subsets for $70\%$ and $30\%$ of the total images respectively.

\item \textit{TRANCOS}~\cite{guerrero2015extremely} is a vehicle crowd counting dataset which is to estimate the number of vehicles in an image of a traffic congestion situation. TRANCOS consists of 1244 traffic jam images with 46796 annotated vehicles. All the collected images contain traffic congestions with a variety of different scenes and viewpoints, covering different lighting conditions, different levels of overlap, and crowdedness.
This dataset is divided into three parts: 403 images for training%the training set
, 420 images for validation, and 421 images for testing.
\end{itemize}

\subsection{Performance Evaluation and Metrics}\label{met:eval}

In evaluating crowd counting networks, the following performance metrics are often used:

\begin{itemize}
    \item \textit{Accuracy:} Accuracy refers to counting accuracy and location accuracy. 
    \begin{itemize}
    \item Counting accuracy is affected by scale variation and isolated clusters of objects~\cite{bai2019crowd}. Scale variation means the same object would appear as a different size in an image due to its perspective and distance from the camera. Besides, an image may have isolated object clusters, and models properly capturing such contextual information usually perform better than others.
    To quantitatively evaluate counting accuracy, Mean Absolute Error (MAE), Mean Squared Error (MSE) and mean Normalized Absolute Error (NAE) are commonly used, defined respectively as:%below:
    %\begin{center}
    %\begin{equation}\small
    $MAE = \frac{1}{N}\sum_{i=1}^{N}|C_{i}-\hat{C}_{i}|$,
    %\end{equation}
    %\begin{equation}\small
    $MSE = \sqrt{\frac{1}{N}\sum_{i=1}^{N}|C_{i}-\hat{C}_{i}|^{2}}$,
    %\end{equation}
    %\begin{equation}\small
    $NAE = \frac{1}{N}\sum_{i=1}^{N}\frac{|C_{i}-\hat{C}_{i}|}{C_{i}}$,
    %\end{equation}
    %\end{center}
    where $N$ is total number of test images, $C_{i}$ the ground truth of the $i$-th image, and $\hat{C}_{i}$  the estimated count. 
    \item Location accuracy is related to the spatial information preserved in the density map. Models with higher quality density map generated usually contains more spatial information for localization tasks.
    \end{itemize}
    
  \item \textit{Quality of density map:} Density map can be evaluated in terms of resolution and visual quality.
    \begin{itemize}
        \item High-resolution density maps usually show better location accuracy and preserve more spatial information for localization tasks (e.g., detection and tracking).
        \item To quantitatively evaluate the visual quality of the generated density maps, Peak Signal-to-Noise Ratio (PSNR) and Structural Similarity in Images (SSIM)~\cite{wang2004image}.
    \end{itemize}
    
    \item \textit{Complexity:} Complexity consists of computational complexity and annotation complexity. 
    \begin{itemize}
    \item Computational complexity is evaluated based on measures such as the number of model parameters, floating-point operations (FLOPs), and inference time.
    \item Annotation complexity, as shown in Table~\ref{tab:4category}, refers to data labeling cost.
      In general, object-level annotation as conducted in the detection-based approach has high complexity.
      Density map estimation requires point-level (head) annotation, which is relatively less costly.
      If unlabeled or synthetic data are used, the complexity can be further reduced.
    \end{itemize}
    \item \textit{Flexibility and robustness:} 
    \begin{itemize}
        \item The flexibility of models is evaluated based on the sensitivity of processing images with arbitrary sizes and the ability to model different kinds of objects (e.g., non-rigid objects).
        \item Robustness refers to distribution shift robustness. It is evaluated in terms of out-of-distribution accuracy, where the test data come from another distribution (w.r.t. the training one).
    \end{itemize}
\end{itemize}

%-----------------------Network-------------------------------------------
%\input{chap/network}

\section{Deep Neural Network Design}\label{sec:net}
Network design is one of the most important parts for density map estimation. In this section, we present the major deep networks for crowd counting: fully convolutional networks (Section~\ref{net:fcn}), encoder-decoder architecture (Section~\ref{net:ed}), multi-column network (Section~\ref{net:multi}),
pyramid structure (Section~\ref{net:pyramid}),
advanced operations (Section~\ref{net:operate}), attention-based model (Section~\ref{net:atten}),
vision transformer (Section~\ref{net:vit}),
and neural architecture search (Section~\ref{net:nas}).
We compare these approaches in Section~\ref{net:summary}, and remark on
some other emerging approaches in Section~\ref{net:others}.

\subsection{Fully Convolutional Network}\label{net:fcn}
\begin{figure*}[t]
	\centering
	\includegraphics[width=1.0\textwidth]{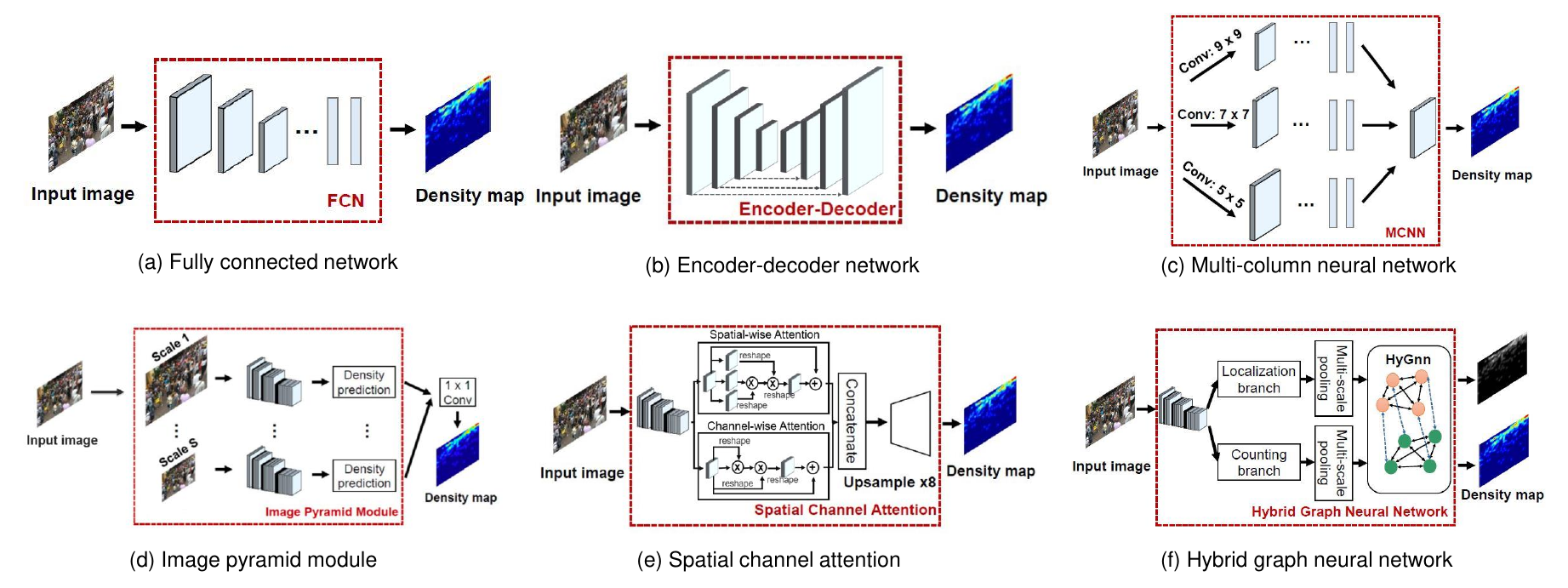}
	\caption{ A summary of the diverse range of network architectures used for deep learning-based single image crowd counting: (a) fully connected network~\cite{marsden2016fully}; (b) encoder-decoder architecture~\cite{cao2018scale}; (c) multi-column network~\cite{zhang2016single}; (d) pyramid structure~\cite{kang2018crowd}; (e) attention-based model~\cite{gao2019scar} ; (f) graph neural network~\cite{luo2020hybrid}. The order of the networks according their presentation in this paper. (Better viewed in the zoom-in mode)}
	\label{fig:network}
\end{figure*}

An early CNN-based density map estimation approach is based on a fully convolutional network (FCN)~\cite{marsden2016fully}, which is modified from the existing CNN architecture (VGG16) and replaces all the fully-connected layers with fully convolutional layers in order to analyze images of arbitrary sizes. As shown in Fig.~\ref{fig:network} (a), FCN learns an end-to-end mapping from an input image to the corresponding density map and 
produces a proportionally sized density map output gave the input image.
The FCN structure is simple but accurate, which has been widely used.

However, the FCN crowd counting method has some limitations.
The resolution of the generated density map is only $1/4$ of the input width and $1/4$ of the input height due to the max pooling operations (extract high-level features but reduce resolutions) in FCN,
which lacks fine details and spatial information for localization tasks, compared with high-resolution density maps.
Besides, the FCN crowd counting model is susceptible to scale variation problems in crowd scene images, which limits its applicability in the general environment.

\subsection{Encoder-Decoder Architecture}\label{net:ed}

The Encoder-decoder model is proposed to align the resolution of the produced density map with the input image.
As shown in Fig.~\ref{fig:network} (b), the encoder-decoder network consists an encoder and a decoder: an encoder network takes the input image and output high-level features, which hold the information and represents the input; a decoder network takes the features from the encoder and generate high-resolution density map.
%In the encoder-decoder architecture,
The encoder gradually downsamples the image resolution with convolutional or pooling layers, and the decoder progressively upsamples the feature maps from the encoder with deconvolutional layers or interpolations. The skip connections are applied on the feature maps from the encoder and decoder respectively.

Some of the deep learning-based
%CNN-based 
crowd counting approaches are following the encoder-decoder structure in recent years (see, for examples, \cite{zhang2019attentional, jiang2019crowd, cao2018scale, liu2018crowd, song2021choose, chen2020scale, thanasutives2021encoder, ding2020crowd}). SANet~\cite{cao2018scale} proposed a novel encoder-decoder network, called scale aggregation Network, which achieves accurate and efficient crowd estimation. The decoder generates high-resolution density maps with a set of transposed convolutions.%}
Furthermore, encoder-decoder based architecture can significantly reduce the number of parameters compared with other architectures due to the downsample operations in the encoder. 
However, such architecture has not addressed the scale variation problem and has not considered the local and global contextual information.

\subsection{Multi-Column Network}\label{net:multi}

Multi-column and pyramid network is
the most prominent models 
in recent crowd counting algorithms to extract the multi-scale features and tackle the scale variation problem
%~\cite{li2019cross, amirgholipour2020pdanet, wu2020counting, wu2019adaptive, liu2019context, yang2020counting, deb2018aggregated}.
~\cite{li2019cross, wu2020counting, wu2019adaptive, liu2019context, yang2020counting, deb2018aggregated, yang2020embedding, wang2021interlayer}.

The multi-column architecture incorporates multi-column architecture with different kernel sizes to extract different scale features in order to achieve accurate counting accuracy such as MCNN~\cite{zhang2016single} and McML~\cite{cheng2019improving}. 
As shown in Fig.~\ref{fig:network} (c), multi-column neural network (MCNN) consists of multiple branches with different kernel sizes (e.g., $5 \times 5$, $7 \times 7$ and $9 \times 9$). The different branches accommodate different receptive fields, thus sensitive to multi-scale features. Finally, the features extracted by different columns are fused together to generate density maps. However, the accommodated scale diversity is restricted by the number of columns.

\subsection{Pyramid Structure}\label{net:pyramid}

Image pyramid and feature pyramid architectures are yet another approach to address scale variations (e.g., AFP~\cite{kang2018crowd}, CP-CNN~\cite{sindagi2017generating}, ~\cite{amirgholipour2020pdanet} and~\cite{xu2022autoscale}), which mainly consists of two subgroups, image pyramid, and feature pyramid pooling. For the image pyramid-based model, as Fig.~\ref{fig:network} (d) shows, different scale of the image pyramid (scale 1, ..., scale S) is feed into an FCN to predict the density map of that scale. Then, the final estimation is produced by adaptive fusing the prediction from different scales. However, this kind of architecture remains a high computational complexity.

Besides, some relevant techniques are usually used together with the multi-column and pyramid networks to enhance the multi-scale feature extraction process such as skip-connections~\cite{sindagi2019pushing, sindagi2019multi, wang2018defense, ding2018deeply, marsden2017resnetcrowd, liu2020crowd} and dense blocks~\cite{oh2020crowd, qiu2019crowd, ma2019atrous, jiang2019learning, idrees2018composition, dai2021dense}.

\subsection{Advanced Convolution Operations}\label{net:operate}

There is a trend to leverage advanced convolutional operations to facilitate accurate crowd counting models and better CNN feature learning~\cite{zhong2022improved, xu2021dilated, huang2020stacked}. 
The deep learning-based
%CNN-based 
single image crowd counting model benefits a lot from the advanced convolution such as dilated and deformable convolution, adaptive dilated convolution, and perspective-guided convolution.
This can replace the traditional convolutional operations in the counting models.
%in the architectures discussed above.

There are four important advanced %convolutional 
operations:

\begin{itemize}
    \item \textit{Dilated convolution} introduces the dilated rate to the convolutional layers, which defines a spacing between the weights of the kernel. Traditional convolutional operation is more focused on extracting local features. For the dilated convolution, %as Fig.~\ref{fig:dilated} shows, 
    three subfigures represent dilated operations with the same kernel size ($3 \times 3$) but different dilated rates (Dilation = $1$, Dilation = $2$, and Dilation = $3$), which enlarges the receptive field without increasing the computational cost and also preserves the resolution of the feature maps. Dilated convolution facilitate real-time applications and is popular in many recent crowd counting models: 
    Dynamic Region Division (DRD)~\cite{he2019dynamic}, Scale Pyramid Network (SPN)~\cite{chen2019scale}, Atrous convolutions spatial pyramid network (ACSPNet)~\cite{ma2019atrous}, DENet~\cite{liu2020denet}, Dilated Convolutional Neural Networks (CSRNet)~\cite{li2018csrnet} and An Aggregated Multicolumn Dilated Convolution Network (AMDCNet)~\cite{deb2018aggregated}. But this kind of operations not consider the multi-scale features and cannot fully capture the non-rigid objects.
    \item \textit{Deformable convolution} is a kind of spatial sampling location augmenting schemes in the modules with additional offsets and learning the offsets from the target tasks, without additional supervision. This can model non-rigid objects with additional learnable offsets. 
    %As shown in Fig.~\ref{fig:deform2}, the first subfigure illustrates the traditional convolutional operation and the second subfigure represents the deformable convolution. The red arrows present the additional offsets. 
    Some recent literatures replace the traditional convolutions with the deformable convolutions and achieves superiors performance: Dilated-Attention-Deformable ConvNet (DADNet)~\cite{guo2019dadnet}, An Attention-injective Deformable Convolutional Network (ADCrowdNet)~\cite{liu2019adcrowdnet}.%, and the Deformation Aggregation Network (DA-Net)~\cite{zou2018net}.
    However, the deformable convolutional operations require high computational complexity.
    \item \textit{Adaptive dilated convolution} is formed to predicts a continuous value of dilation rate for each location in order to effectively match the scale variation at different locations, which is better than fixed and discrete dilate rates.
    ADSCNet~\cite{bai2020adaptive} is formulated based on adaptive dilated convolution, which is also able to preserve the strong consistency between the density and feature of each location.
    \item \textit{Perspective-guided convolution} aims to tackle the continuous scale variation issue with perspective information. The perspective information contains instance information between camera and a scene, which is a reasonable prior for people scale estimation. Concretely, the perspective information functions are leveraged to guide the spatially variant smoothing of feature maps before feeding to the successive convolutions. PGCNet~\cite{yan2019perspective} is built by stacking multiple Perspective-guided convolutions (PGC) blocks based on a CNN backbone, which is a single-column CNN target to tackle the scale variation issues with a moderate increase in computation.
\end{itemize}

\begin{table*}
	\centering
	\caption{Comparisons of network design considerations for crowd counting. \textbf{Computational complexity} is evaluated based on the number of model parameters. The representative schemes of each network design category are analyzed thoroughly in terms of \textbf{advantages} and \textbf{limitations}.}
	\begin{tabular}{|c|c|c|c|c|c|}
		\hline
		\multicolumn{2}{|c|}{\textbf{Category}} &\textbf{\tabincell{c}{Representative\\Scheme}} &\textbf{Advantages} &\textbf{\tabincell{c}{Computational \\Complexity}}
		&\textbf{Limitations}\\
		\hline
		\multicolumn{2}{|c|}{
		\tabincell{c}{Fully convolution \\neural networks}}
		&FCN~\cite{marsden2016fully}
		&\tabincell{c}{Can analyze \\images of \\arbitrary size}
		&Low
		&\tabincell{c}{Low-resolution \\density maps}
		\\
		\hline
		\multicolumn{2}{|c|}{
		\tabincell{c}{Encoder-decoder \\architecture}}
		&SANet~\cite{cao2018scale} 
		&\tabincell{c}{Able to generate \\high-resolution \\density maps} 
		&\tabincell{c}{Low(0.9M)}%0.91
		&\tabincell{c}{Not consider \\scale variation} \\
		\hline
		\multicolumn{2}{|c|}{
		\tabincell{c}{Multi-column \\architecture}}
		&{MCNN~\cite{zhang2016single}} 
		&{\tabincell{c}{Extract multi-scale \\features with \\multi-column \\architecture}}
		&{\tabincell{c}{Low(0.1M)}}%0.13
		&{\tabincell{c}{The scale diversity \\is restricted by \\the number \\of columns}} \\
		\hline
		\multicolumn{2}{|c|}{
		{
		\tabincell{c}{Pyramid \\architecture}}}
		&{CP-CNN~\cite{sindagi2017generating}}
		&{\tabincell{c}{Extract multi-scale \\features with \\pyramid architecture}}
		&{\tabincell{c}{High(68.4M)}}
		&{\tabincell{c}{High computational\\ complexity}}
		\\
		\hline
		\multirow{4}{*}{\tabincell{c}{{ Advanced} \\{ convolution} \\{ operations}}} &
		\multicolumn{1}{|c|}{
		\tabincell{c}{Dilated \\convolution \\operations}}
		&CSRNet~\cite{li2018csrnet}
		&\tabincell{c}{Enlarge receptive \\field without \\increase the \\computational cost}
		&\tabincell{c}{Medium(16.3M)}%16.26
		&\tabincell{c}{Not consider the \\non-rigid objects}\\
		\cline{2-6}
		~ &
		\multicolumn{1}{|c|}{
		\tabincell{c}{Deformable \\convolution \\operations}}
		&ADCrowdNet~\cite{liu2019adcrowdnet}
		&\tabincell{c}{Learnable additional \\offsets for \\better modeling \\non-rigid objects}
		&High
		&\tabincell{c}{High computational\\ complexity}
		\\
		\cline{2-6}
		~ &
		\multicolumn{1}{|c|}{
		{
		\tabincell{c}{ Adaptive \\dilated \\convolution}}
		}
		&{ ADSCNet~\cite{bai2020adaptive}}
		&{\tabincell{c}{Learn continuous\\dilation rate}}
		&{ Medium}
		&
		{
		\tabincell{c}{Not flexible\\to non-rigid\\objects}
		}
		\\
		\cline{2-6}
		~ &
		\multicolumn{1}{|c|}{
		{
		\tabincell{c}{Perspective-\\guided \\convolution}}
		}
		&{ PGCNet~\cite{yan2019perspective}}
		&{\tabincell{c}{Perspective\\information\\facilitate people\\scale estimation}
		}
		&{Medium}
		&
		{
		\tabincell{c}{Requires additional\\ perspective\\information}
		}
		\\
		\hline
		\multicolumn{2}{|c|}{
		\tabincell{c}{Attention-based \\Model}}
		&SCAR~\cite{gao2019scar}
		&\tabincell{c}{Capture local \\and global \\contextual \\information}
		&Medium
		&\tabincell{c}{Rely on pixel-wise \\loss function}\\
		\hline
		\multicolumn{2}{|c|}{\tabincell{c}{Vision Transformer}} &{ TransCrowd~\cite{liang2022transcrowd}}&{\tabincell{c}{Able to model\\long-range context\\ information}}&{ Medium}&{\tabincell{c}{Computational\\expensive}} \\		
		\hline
		\multicolumn{2}{|c|}{{\tabincell{c}{Neural Architecture\\Search}}} &{ NAS-Count~\cite{hu2020count}}&{\tabincell{c}{Automate crowd\\counting model\\design}}&{ Medium}&{\tabincell{c}{Computational\\expensive}} \\	
		\hline		
	\end{tabular}	
	\label{tab:netdesign-ana}
\end{table*}

\subsection{Attention-based Model}\label{net:atten}
Attention mechanisms can be roughly divided into two subgroups: hard attention and soft attention~\cite{lin2022boosting, wang2022hybrid, wang2022crowd, wu2021cranet, sajid2022towards, jiang2020attention, duan2020sofa, hou2020bba, chen2020crowd, chen2020relevant}. Such mechanisms have been explicitly explored in recent years, and we summarize several recent algorithms applied with the attention mechanism: AFPNet~\cite{kang2018crowd}, MRA-CNN~\cite{zhang2019multi}, SAAN~\cite{hossain2019crowd}, DADNet~\cite{guo2019dadnet}, Relational Attention Network~\cite{zhang2019relational}, Hierarchical Scale Recalibration Network~\cite{zou2020crowd}, ACM-CNN~\cite{zou2019attend}, HA-CNN~\cite{sindagi2019ha}, Shallow Feature-based Dense Attention Network~\cite{miao2020shallow} and Multi-supervised Parallel Network~\cite{wei2020mspnet}.

SCAR~\cite{gao2019scar} is one of the typical models to make use of attention schemes. SCAR proposes a spatial- /channel-wise attention regression module for crowd counting.
As shown in Fig.~\ref{fig:network} (e), the top half branch (spatial-wise attention) captures large-range contextual information and the change of density distribution, which the output feature map is weighted sum of attention map and original local feature map.
The bottom half branch shows the channel-wise attention, which leverages both local and global contextual information for crowd counting. 
The features extracted by these two branches are late fused by concatenation and upsample post-processing to generate density maps. However, most of the methods discussed above are relying on pixel-wise loss functions for optimizing the model. We will discuss%some 
advanced loss functions to better capture spatial correlations between pixels and to generate high-quality density maps.

\begin{table*}
	\centering
	%\doublespacing
	\caption{Quantitative comparisons of different network design considerations on widely used crowd counting datasets. The counting accuracy is evaluated based on \textbf{MAE} and \textbf{MSE}. The visual quality of the generated density maps is evaluated based on \textbf{PSNR} and \textbf{SSIM}. \textbf{ST PartA} and \textbf{ST partB} denotes ShanghaiTech A and ShanghaiTech B dataset~\cite{zhang2016single}, respectively.}
	%\scalebox{0.5}{
	\begin{tabular}{|c|c|c|c|c|c|c|c|c|c|c|}
		\hline
		\multicolumn{3}{|c|}{\textbf{Representative Schemes}} & \multicolumn{4}{|c|}{\textbf{ST PartA}} & \multicolumn{2}{|c|}{\textbf{ST PartB}} &
		\multicolumn{2}{|c|}{\textbf{UCF\_CC\_50}} %&
		%\multicolumn{2}{|c|}{\textbf{UCSD}} 		
		\\
		\hline
		\textbf{\tabincell{c}{Methods}} &\textbf{Year} &\textbf{Column} &\textbf{MAE}  &\textbf{MSE} 
		&\textbf{PSNR} &\textbf{SSIM}
		&\textbf{MAE}  &\textbf{MSE} &\textbf{MAE}  &\textbf{MSE}  %&\textbf{MAE}  &\textbf{MSE} 
		\\
		\hline
		FCN~\cite{marsden2016fully} &2016 &Single
		&126.5 &173.5 & - & - &23.8 &33.1
		&338.6 &424.5 %& - & - %& - & - 
		\\
		MCNN~\cite{zhang2016single} &2016 &Multi   
		&110.2  &173.2 &21.40 &0.52 &26.4  &41.3 
		&377.6  &509.1 %&1.07 &1.35 %&277   &426   
		\\
		\hline
		CP-CNN~\cite{sindagi2017generating}&2017&Multi &73.6  &106.4 &21.72 &0.72 &20.1  &30.1
		&295.8  &320.9 %& - & - %& - & -  
		\\
		\hline
		SANet~\cite{cao2018scale} &2018 &Single 
		&67.0  &104.5 & - & - &8.4   &13.6
		&258.4 &334.9 %&1.02 &1.29 %& -  & -   
		\\
		CSRNet~\cite{li2018csrnet} &2018 &Single   
		&68.2   &115.0 &23.79 &0.76 &10.6 &16.0 
		&266.1  &397.5 %&1.16  &1.47 %& -   & -  
		\\
		\hline
		ADCrowd~\cite{liu2019adcrowdnet}&2019&Single &63.2  &98.9 &24.48 &0.88 &8.2 &15.7
		&266.4 &358.0  %&1.10 &1.42 %& - & - 
		\\
		{ PGCNet~\cite{yan2019perspective} } &{2019} &{Single} 
		&{57.0} &{86.0} & {-} & {-} &{8.8} &{13.7}
		&{244.6} &{361.2} %& {-} & {-}
		\\
		SCAR~\cite{gao2019scar} &2019 &Double 
		&66.3 &114.1 &23.93  &0.81  &9.5 &15.2
		&259.0 &374.0 %& - & - %& - & - 
		\\
		\hline
		{ADSCNet~\cite{bai2020adaptive} } &{2020} &{Single} 
		&{60.7} &{100.6} & {-} & {-} &{6.4} &{11.3}
		&{198.4} &{267.3} %& {-} & {-}
		\\
		{ MS-GAN~\cite{zhou2020adversarial}} &{2020} &{Single} 
		&{-} &{-} & {-} & {-} &{18.7} &{30.5}
		&{345.7} &{418.3} %&{1.78} & {3.03} %&100.8 &185.3 
		\\	
     	HyGnn~\cite{luo2020hybrid} &2020 &Double 
		&60.2 &94.5 & - & - &7.5 &12.7
		&184.4 &270.1 %& - & - %&100.8 &185.3 
		\\
		\hline
		{TransCrowd~\cite{liang2022transcrowd}} &{2021} &{Single}
		&{66.1} &{105.1} &{-} &{-} &{9.3} &{16.1} &{-} &{-} 
		%&{-} &{-}
		\\
		{NAS-Count~\cite{hu2020count}} &{2021} &{Single}
		&{56.7} &{93.4} &{-} &{-} &{6.7} &{10.2} 
		&{208.4} &{297.3} %&{-} &{-}
		\\
		\hline
		STNet~\cite{wang2022stnet} &2022 &Single
		&52.9 &83.6 & - & - & 6.3 & 10.3 
		& 162.0 & 230.4 %& - & -
		\\
		\hline
	\end{tabular}
	%}
	\label{tab:netdesign-acc}
\end{table*}

\subsection{Vision Transformer}\label{net:vit}
The mainstream crowd estimation approaches usually leverage the convolution neural network to extract features and significant progress has been achieved by incorporating larger context information into CNNs, which indicates that long-range context is essential. The self-attention mechanisms of transformers, which explicitly model all pairwise interactions between elements in a sequence, which is particularly suitable to extract the semantic crowd information.

TransCrowd~\cite{liang2022transcrowd} proposes two different kinds of approaches for single image crowd counting: TransCrowd-Token and TransCrowd-GAP, which can generate reasonable attention weight and achieve high counting performance.

\subsection{Neural Architecture Search}\label{net:nas}
Most of the recent advances in counting network design are based on hand-designed neural networks, which require large design efforts and strong domain knowledge. To extract multi-level features, convolutions with various receptive fields are designed by hand.
Recently, automatic and lightweight network design has drawn much attention. Automated Machine Learning %(AutoML) 
and Neural Architecture Search (NAS) techniques can be used to automatically design effective and efficient crowd counting architectures~\cite{wang2022eccnas}. And the NAS-based approach is able to automatically discover the task-specific multi-scale crowd estimation models.

NAS-Count~\cite{hu2020count} automates the design of crowd counting models with NAS and proposes an end-to-end searched encoder-decoder architecture, where multi-scale features can be leveraged to tackle the scale variation problem. 
%It should notice that 
The first attempt in NAS needs hundreds of GPUs to run. However, NAS-Count leverage a differential one-shot search strategy to achieve fast search speed, where network parameters and architecture parameters are jointly optimized via gradient descent.
In addition, NAS-Count is enabled by the compositional nature of CNN and is guided by task-specific search space and strategies. The architectures searched by the counting-oriented NAS framework achieve superior performance without
demanding expert-involvement.

\subsection{Comparisons}\label{net:summary}

We compare the different networks discussed above
in Table~\ref{tab:netdesign-ana}, and present
their performance on three challenging crowd counting datasets in Table~\ref{tab:netdesign-acc}. We also provide a comprehensive performance analysis of state-of-the-art crowd counting approaches in Table~\ref{tab:quan}.
By analyzing the data, we find some intriguing observations. 

As Tables~\ref{tab:netdesign-ana} and~\ref{tab:netdesign-acc} show, SANet achieves better counting performance on datasets with different crowd levels, compared with FCN. The generated density maps of FCN are only $1/4 \times 1/4$ of the original input image, which SANet is able to generate high-resolution density maps. The computational complexity for both the FCN and SANet is low (e.g., 0.91M for SANet), which indicates that the encoder-decoder architecture is lightweight.

MCNN and CP-CNN consider scale variation problem, which is able to capture multi-scale features. MCNN extracts multi-scale features with multi-column architecture and CP-CNN extracts multi-scale features with pyramid architecture. CP-CNN achieves better counting accuracy and visual quality than MCNN, while for the computational complexity, the number of parameters for CP-CNN (68.4M) is much larger than MCNN (0.13M). This further demonstrates the effectiveness of multi-column architecture and pyramid architecture, while image pyramid architecture (e.g., CP-CNN) is of high computational complexity.

CSRNet and ADCrowdNet achieve better counting accuracy and visual quality than MCNN and CP-CNN on most of the datasets. CSRNet relies on dilated convolutional operations, which enlarge the receptive field without increase the computational cost. ADCrowdNet incorporates deformable convolutional operations, which are based on learnable additional offsets for better modeling non-rigid objects such as people. In addition, ADCrowdNet achieves better counting accuracy and visual quality than CSRNet but requires higher computational complexity.

SCAR shows better counting accuracy and visual quality than MCNN and CP-CNN, which is able to capture local and global contextual information based on spatial-wise attention and channel-wise attention schemes. The experimental results confirm the effectiveness of attention mechanism variations for crowd counting. HyGnn shows good counting performance on different crowd counting datasets, which demonstrates the effectiveness of graph-based models to distill rich relations among multi-scale features for crowd counting.

The multi-path encoder-decoder network searched by NAS-Count demonstrates better performance than tedious hand-designing crowd counting models on four challenging datasets, which achieves a multi-scale model automatically without strong domain knowledge. This clearly demonstrates the potential to automatically design effective and efficient crowd counting architectures.

\subsection{Others}\label{net:others}
There are also some other emerging network designs for crowd counting, discussed below:

\begin{itemize}
\item \textit{Generative Adversarial Networks}
Generative Adversarial Networks (GAN) has been applied to a wide range of tasks in computer vision, and also have been adopted to crowd counting tasks such as GAN-MTR~\cite{olmschenk2018crowd}, MS-GAN~\cite{yang2018multi},~\cite{zhou2020adversarial}, ACSCP~\cite{ma2019atrous} and CODA~\cite{li2019coda}.
Generative adversarial networks can be used to improve the visual quality of the generated density maps, but usually degrades counting accuracy.
For example, MS-GAN~\cite{yang2018multi, zhou2020adversarial} proposed multi-scale GAN, which incorporates the inception module in the generation part.
This paper investigated GAN as an effective solution to the crowd counting problem, to generate high-quality crowd density maps of arbitrary crowd density scenes.
Besides, Adversarial Cross-Scale Consistency Pursuit (ACSCP)~\cite{ma2019atrous} designed a novel scale-consistency regularizer that enforces that the sum up of the crowd counts from local patches. The authors further boosted density estimation performance by further exploring the collaboration between both objectives. 
    \item \textit{Graph neural networks} based method distills rich relations among multi-scale features for crowd counting. As shown in Fig.~\ref{fig:network} (f), 
    %As shown in Fig.~\ref{fig:hygnn}, 
    %Hybrid Graph Neural Networks
    HyGnn~\cite{luo2020hybrid} exploits useful information from the auxiliary task (localization branch). The HyGnn module in the red box jointly represents the task-specific feature maps of different scales as nodes, multi-scale relations as edges, counting, and localization relations as edges, which distilled rich relations between the nodes to obtain more powerful representations, leading to robust and accurate results.
    \item \textit{Recurrent neural networks} based Deep Recurrent Spatial-Aware Network (DSRNet)~\cite{liu2018crowd} utilize a learnable spatial transform module with a region-wise refinement process to adaptively enlarge the varied scales coverage. Researchers in~\cite{shang2016end} decoded the features into local counts using an LSTM decoder, finally predicts the image global count. The local counts and global count are all learning targets.
    \item \textit{Prior-guided modules} help enhancing counting performance, as discussed in recent literature~\cite{lian2021locating, rong2021coarse, yan2021crowd, yang2020reverse,sajid2021multi, pan2020attention, mo2020background, zhao2019scale, jiang2019mask, wang2019removing}.
    %\item \textit{Multi-stage density map regression network}
    Multi-stage density map regression network is a scale-aware convolutional neural network (MMNet)~\cite{dong2020crowd}, which not only captures multi-scale features generated by various sizes of filters but also integrates multi-scale features generated by different stages to handle scale variation problems.
    \item \textit{Local counting network} proposes an adaptive mixture regression framework~\cite{liu2020adaptive} in a coarse-to-fine manner to improve counting accuracy, which fully utilizes the context and multi-scale information from different convolutional features. Besides, local counting networks perform more precise counting regression on local patches of images.
    %\item \textit{Combining with detection} 
    \item \textit{Multi-model fusion} is another class of techniques for crowd counting~\cite{sajid2021audio, liu2021cross, tang2022tafnet}. Recently, most of the current works for crowd counting with state-of-the-art performance are density-map estimation-based approaches. Some researchers tried to improve the existing framework with both point and box annotation such as LCFCN~\cite{laradji2018blobs}, PSDDN~\cite{liu2019point}, BSAD~\cite{huang2017body}, DecideNet~\cite{liu2018decidenet} and DRD~\cite{he2019dynamic}.  DecideNet~\cite{liu2018decidenet} is one of the typical methods, which proposed a separate decide subnet to combine detection and density estimation. Combining detection with density map estimation usually utilizes detection for the low crowd and density estimation for the high crowd. However, these kinds of methods require high computational complexity and high annotation complexity.
\end{itemize}

%------------------------------Loss---------------------------------------
%\input{chap/loss}
\section{Loss Function}\label{sec:loss}
The loss functions are used to optimize the model. Early works usually adopt the pixel-wise Euclidean loss (Section~\ref{loss:l2}), later different advanced loss functions are utilized for better density estimation.
In this section, we discuss some recent advances on loss functions for crowd counting: SSIM loss (Section~\ref{loss:ssim}), 
and multi-task learning (Section~\ref{loss:multi}).
We compare them in Section~\ref{loss:summary} and present some 
other emerging considerations in Section~\ref{loss:ensemble}.

\subsection{Euclidean Loss}\label{loss:l2}
Most of the early crowd counting approaches use Euclidean loss to optimize the models. The Euclidean loss is a pixel-wise estimation error:
\begin{equation}
L_{E}=\frac{1}{N} \left ||F(x_{i};\theta)-y_{i} \right ||_{2}^{2},
\end{equation}
where $\theta$ indicates the model parameters, N means the number of pixels, $x_{i}$ denotes the input image, and $y_{i}$ is ground truth and $F(x_{i};\theta)$ is the generated density map. The total crowd counting result can be summarized over the estimated crowd density map.
The pixel-wise L2 loss is a flexible and widely used loss function for crowd counting. However, this pixel-wise loss does not take local and global contextual information as well as the visual quality of the generated density maps into account. Thus, this kind of loss function cannot produce satisfactory high-quality density maps and highly accurate crowd estimation.

\subsection{SSIM Loss}\label{loss:ssim}
Some variants of structure similarity (SSIM) loss are proposed for crowd counting to force the network to learn the local correlation within regions of various sizes, thereby producing locally consistent estimation results such as SSIM loss~\cite{cao2018scale}, multi-scale SSIM loss~\cite{qiu2019crowd}, DMS-SSIM loss~\cite{liu2019crowd} and DMSSIM loss~\cite{li2019cross}.
%The SSIM index can be calculated point by point as Equation~\ref{ssim}
Then the local pattern consistency can be formulated as:
\begin{equation}
%\centering
L_{s} = 1 - \frac{1}{N}\sum_{x}SSIM(x).
\end{equation}
The pixel-wise Euclidean loss usually assumes that adjacent pixels are independent and ignores the local correlation in the density maps, the Euclidean loss can be fused with the SSIM loss to leverage local correlations among pixels for generating high-quality density maps and accurate crowd estimation. 

For example, the Cross-Level Parallel Network~\cite{li2019cross} fused the difference of mean structural similarity index (DMSSIM) with the MSE loss to optimize the module. Besides, Multi-View Scale Aggregation Networks~\cite{qiu2019crowd} proposed a multi-scale SSIM for multi-view crowd counting.
However, SSIM loss is hard to learn local correlations with a large spectrum of varied scales.

\subsection{Multi-task Learning}\label{loss:multi}

The main task of crowd counting is the total counting accuracy, thus the direct global count constraints may benefit the counting accuracy.
The headcount loss can be defined as:
\begin{equation}
L_c = \frac{1}{N}\sum_{i=1}^{N}||\frac{F_c(x_{i};\theta) - y_{i}}{y_{i} + 1}||,
\end{equation}
where $F_c(x_{i};\theta)$ is the estimated head count, and $y_{i}$ is the ground truth head count.
Then the total loss function is formulated as follow:
\begin{equation}
L_{total} = L_{E} + \alpha L_{c},
\end{equation}
where $\alpha$ is the weight to balance the pixel-wise Euclidean loss and the total head counting loss.
BL~\cite{ma2019bayesian} stated that the original GT density map is imperfect due to occlusions, perspective effects, variations in object shapes and proposed Bayesian loss to constructs a density contribution probability model from the point annotations and addressed the above issues. The proposed Bayesian loss adopted more reliable supervision on the count expectation at each annotated point.

SaCNN~\cite{zhang2018crowd} proposed to combine density map loss with the relative count loss. The relative count loss helps to reduce the variance of the prediction errors and improve the network generalization on very sparse crowd scenes. CFF~\cite{shi2019counting} fused segmentation map loss, density map loss and global density loss. Plug-and-Play Rescaling~\cite{sajid2020plug} combined regression loss with classification loss. Shallow Feature-based Dense Attention Network~\cite{miao2020shallow} proposed to use MSE loss with counting loss and stated that counting loss not only accelerates the convergence but also improves the counting accuracy.
Multi-supervised Parallel Network~\cite{wei2020mspnet} combined MSE loss, cross-entropy loss, and L1 loss.
Besides, there is also some paper to use a kind of combination loss to enforce similarities in local coherence and spatial correlation between maps~\cite{jiang2019crowd},~\cite{ranjan2018iterative}~\cite{idrees2018composition}.
Multi-task learning based framework is widely used in recent papers~\cite{tan2019crowd},~\cite{liu2019recurrent},~\cite{han2020focus},~\cite{kumar2019mtcnet},~\cite{gao2019pcc},~\cite{zhang2019multi},~\cite{sindagi2017cnn}. However, this kind of framework is sensitive to hyper-parameters.

\subsection{Comparisons}\label{loss:summary}

We summarize the advantages and limitations of the above loss functions
in Table~\ref{tab:loss-ana}. We compare in Table~\ref{tab:loss-acc} the performance of several state-of-the-arts with different loss functions.

\begin{table*}
	\centering
	%\doublespacing
	\caption{ Comparisons of recent advanced loss functions for crowd counting. The property of representative schemes for each loss functions category are summarized based on \textbf{advantages} and \textbf{limitations}.}
	\begin{tabular}{|c|c|c|c|}
		\hline
		\textbf{Category} &\textbf{Representative Scheme} &\textbf{Advantages} &\textbf{Limitations}\\
		\hline
		Euclidean loss 
		&CSRNet~\cite{li2018csrnet}
		&\tabincell{c}{Flexible; widely used}
		&\tabincell{c}{Not consider context\\information and visual quality}
		\\
		\hline
		SSIM loss
		&SANet~\cite{cao2018scale}
		&\tabincell{c}{Variants of structural similarity\\loss to learn local correlation}
		&\tabincell{c}{Hard to learn the local\\correlation with various scales}
		\\
		\hline
		{Multi-task learning}
		&MSPNet~\cite{wei2020mspnet}
		&
		{
		\tabincell{c}{Varied and flexible\\to fuse different constrains}
		}
		&
		{
		\tabincell{c}{Sensitive to hyper-parameters}
		}
		\\
		\hline
		Others
		&S-DCNet~\cite{xiong2019open}
		&\tabincell{c}{Efficient divide\\and conquer manner}
		&\tabincell{c}{Computational expensive}
		\\
		\hline
	\end{tabular}	
	\label{tab:loss-ana}
\end{table*}

\begin{table*}
	\centering
	%\doublespacing
	\caption{
	Comparisons of state-of-the-art crowd counting approaches with different loss functions. \textbf{Multi scale} is the multi-scale design considerations; \textbf{Dilated} is dilated convolutions; \textbf{Deform} is the deformable convolutions; \textbf{Atten} represents the attention-based scheme. \textbf{ST-A} denotes ShanghaiTech A dataset~\cite{zhang2016single} and \textbf{ST-B} denotes ShanghaiTech B dataset~\cite{zhang2016single}.}
	\begin{tabular}{|c|c|c|c|c|c|c|c|c|c|}
		\hline
		\textbf{Scheme} %&\textbf{Year}
		&\textbf{\tabincell{c}{Multi\\scale}}  &\textbf{Dilated} 
		&\textbf{Deform} 
		&\textbf{Atten} 
		&\textbf{\tabincell{c}{Loss function}}
		&\textbf{\tabincell{c}{ST-A \\MAE}} &\textbf{\tabincell{c}{ST-A \\MSE}} &\textbf{\tabincell{c}{ST-B \\MAE}} &\textbf{\tabincell{c}{ST-B \\MSE}}\\
		\hline
		CSRNet~\cite{li2018csrnet} %&2018 
		& &$\surd$ & & &\tabincell{c}{Euclidean loss} 
		&68.2 &115.0 &10.6 &16.0 \\
		\hline
		ADCrowd~\cite{liu2019adcrowdnet}%&2019  
		& & &$\surd$ &$\surd$ &\tabincell{c}{Euclidean loss} 
		&63.2 &98.9 &8.2 &15.7 \\
		\hline
		DSSINet~\cite{liu2019crowd} %&2019
		&$\surd$ &$\surd$ & & &SSIM Loss
		&60.63 &96.04 &6.8 &10.3 \\
		\hline
		S-DCNet~\cite{xiong2019open} %&2019
		&$\surd$ & & &  &Divide-conquer
		&58.3 &95.0 &6.7 &10.7\\
		\hline
		HA-CCN~\cite{sindagi2019ha} %&2020
		& & & &$\surd$ &\tabincell{c}{MSE loss with \\global counting}
		&62.9 &94.9 &8.1 &13.4 \\
		\hline
		GLoss~\cite{wan2021generalized} %&2021 
		& & & & &\tabincell{c}{Unbalanced optimal\\transport loss}
		&61.3 &95.4 &7.3 &11.7 \\
		\hline
	\end{tabular}	
	\label{tab:loss-acc}
\end{table*}

CSRNet and ADCrowdNet are based on the same Euclidean loss but with different deep neural network designs and show different counting accuracy, which shows that the Euclidean loss is flexible and widely used in the early approaches. However, the Euclidean loss lacks contextual information and ignores the local correlation among pixels in the density maps.

The DSSINet achieves better performance than CSRNet and ADCrowdNet on different crowd counting datasets. 
These variants of structural similarity loss show counting improvements based on utilizing local correlation. However, these kinds of methods suffer in the situation of a large spectrum of various scales. 

As Table~\ref{tab:loss-acc} shows, DSSINet (SSIM loss) achieves better counting accuracy than ACSCP (Adversarial loss) with similar network design considerations (i.e., multi-scale scheme and dilated convolutional operations). The poor performance of ACSCP on ShanghaiTech A \& B may probably be due to the adversarial loss. This further demonstrates that adversarial loss can help to generate high-quality density maps but may sacrifice counting accuracy. 

HA-CNN shows better performance than ADCrowdNet even without deformable convolutional operations on two different crowd counting datasets. This demonstrates that multi-task learning with global counting constrain can work well in highly crowded scenes even without some advanced network operations. S-DCNet also achieves satisfactory counting accuracy on different crowd counting datasets, which confirms the effectiveness of the divide and conquer manner but is computationally expensive.

\subsection{Others}\label{loss:ensemble}

There are some other loss optimization strategies to enhance crowd counting tasks~\cite{liu2020weighing, wang2021uniformity, wang2020density, jiang2020density, servadei2022label}.
CNN-Boosting~\cite{walach2016learning} employed CNNs and incorporate two significant improvements: layered boosting and selective sampling. DAL-SVR~\cite{wei2019boosting} boosted deep attribute learning via support vector regression for fast-moving crowd counting.
The paper learned superpixel segmentation-fast moving segmentation-feature extraction-motion features/appearance features/sift feature-features aggregation by PCA-regression learning SVR-data fusion and deeply learning cumulative attribute.
D-ConvNet~\cite{shi2018crowd} used seep negative correlation learning, which is a successful ensemble learning technique for crowd counting. The authors extended D-ConvNet in~\cite{zhang2019nonlinear}, which proposed to regress via an efficient divide and conquer manner.
D-ConvNet has been shown to work well for non-deep regression problems. Without extra parameters, the method controls the bias-variance-covariance trade-off systematically and usually yields a deep regression ensemble where each base model is both accurate and diversified. However, the whole framework is computationally expensive.

S-DCNet~\cite{xiong2019open} designed a multi-stage spatial divide and conquer network.
The collected images and labeled count values are limited in reality for crowd counting, which means that only a small closed set is observed. A dense region can always be divide until sub-region counts are within the previously observed closed set. S-DCNet only learns from a closed set but can generalize well to open-set scenarios. And avoid repeatedly computing sub-region convolutional features, this method is also efficient.

%-----------------------------Supervision--------------------------------
%\input{chap/supervision}
\section{Supervisory Signal}\label{sec:data}

In this section, we discuss different supervisory signals for crowd counting: fully supervised learning (Section~\ref{data:gt}), weakly supervised and semi-supervised learning (Section~\ref{data:weakly}), unsupervised and self-supervised learning (Section~\ref{data:unlabel}), and automatic labeling through synthetic data (Section~\ref{data:aug}). We evaluate and compare them in Section~\ref{data:summary}.

\subsection{Fully Supervised Learning} \label{data:gt}

In the fully supervised crowd counting paradigm, the model is hard to optimize if we utilize the original discrete point-wise annotation maps as ground truth~\cite{liang2021focal, gao2020learning, zand2022multiscale, louedec2021gaussian, liu2021bipartite, abousamra2021localization, wang2021dense, zhou2021locality, wan2021fine, sam2020locate}.
There are also some recent works study the problem of counting from scalar representations~\cite{wang2022crowdmlp, song2021rethinking, lin2021direct, ma2021learning}.
The continuous ground truth density map is usually generated from the original point-wise annotations via different ground truth generation methods such as applying an adaptive Gaussian kernel for each head annotation, which is important for accurate crowd estimation~\cite{wan2020kernel}. 
The fixed kernel or adaptive Gaussian kernel are widely used approaches to prepossess the original annotation and get the ground truth for density estimation and crowd counting~\cite{li2018csrnet}. The geometry-adaptive kernel is defined as follows:
\begin{equation}
F(x) = \sum_{i=1}^{N}\delta(x-x_i) \times G_{\sigma_i}(x), with \; \sigma_i = \beta\bar{d_i},
\end{equation}
where x denotes the pixel position in an image. For each target object, $x_i$ in the ground truth, which is presented with a delta function $\delta(x-x_i)$. The ground truth density map $F(x)$ is generated by convolving $\delta(x-x_i)$ with a normalized Gaussian kernel based on parameter $\sigma_i$.
And $\bar{d_i}$ shows the average distance of the k nearest neighbors.

\begin{figure*}[t]
	\centering
	\includegraphics[width=0.85\textwidth]{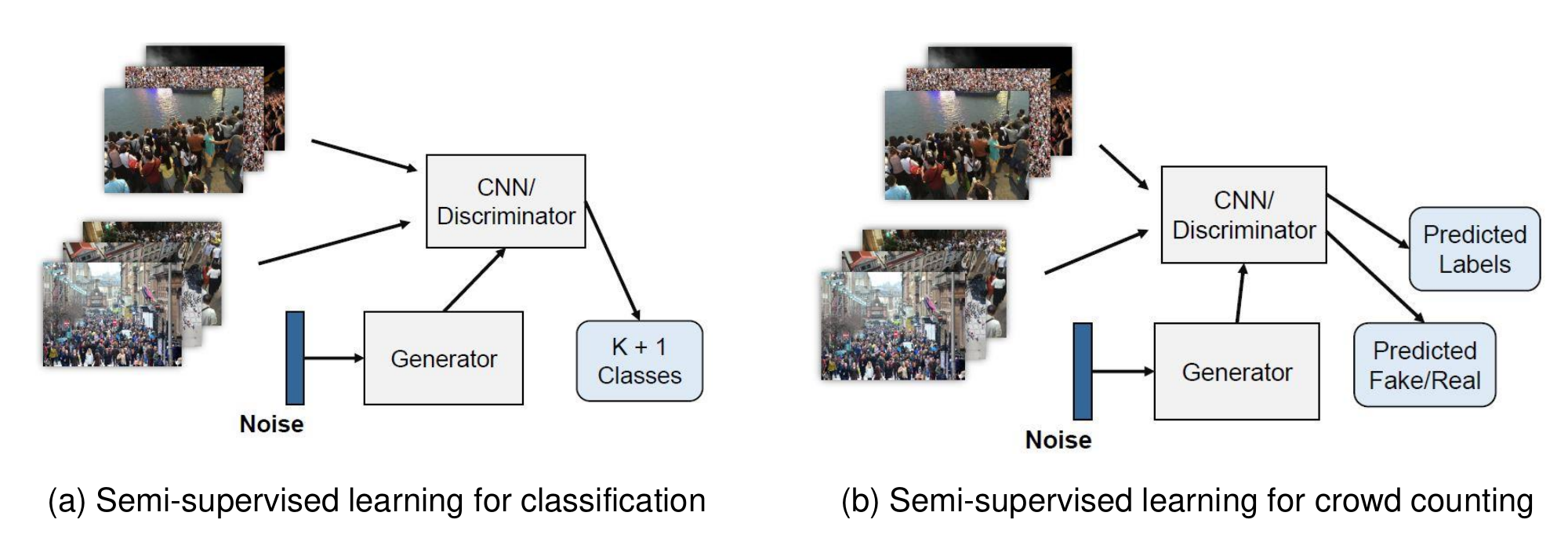}
	\caption{The workflow of the original semi-supervised learning for classification problem (Figure a) and semi-supervised learning for single image crowd counting (Figure b)~\cite{olmschenk2019dense}.}
	\label{fig:semi}
\end{figure*}

GP~\cite{von2016gaussian} devises a Bayesian model that places a Gaussian process before a latent function whose square is the count density.
Compared to different annotation methods concerning their difficulty for the annotator: dots or bounding box in all objects, GP is better in terms of accuracy and labeling effort.
Besides, there are some recent advances to use a learned kernel to improve the prepossessing step and proposed an adaptive density map generator~\cite{wan2019adaptive}.

DM-Count~\cite{wang2020distribution} optimizes the network directly on the dot map, which can be considered as a special type of density map with $1 \times 1$ Gaussian blur. Most existing %crowd counting 
methods need to use an adaptive or fixed Gaussian to smooth each annotated dot or to estimate the likelihood of every pixel given the annotated point. DM-Count directly optimizes the original annotation and shows its generation error bound is tighter than that of Gaussian smoothed methods.

\subsection{Weakly Supervised and Semi-supervised Learning}\label{data:weakly}
Recently, a number of works have emerged to make use of weakly labeled data for crowd counting~\cite{meng2021spatial, wang2021self, khaki2021deepcorn, sindagi2020learning, liu2020semi, zhao2020active, yang2020weakly, kong2020weakly, zhou2018crowd} and the problem of learning from noisy annotations~\cite{wan2020modeling, li2021learning}. The original annotation process for crowd counting via density map estimation is point-level annotation, which is labor-intensive, 
HA-CCN~\cite{sindagi2019ha} proposed a weakly supervised learning setup and leveraged the image-level labels instead of the densely point-wise annotation process to reduce label effort. 
%As shown in Fig.~\ref{fig:wsl}, 
As shown in Fig.~\ref{fig:flow} (b), 
the first column is the original image, the second column is the labor-intensive dense (head) annotation, the third column is the ground truth density maps, and the last column is the image-level weak annotation, which is used in the weakly supervised learning setting. This clearly shows that leveraging weakly labeled data (the last column) can largely reduce the annotation complexity compared with fully point-wise annotation (the second column). Besides, Scale-Recursive Network (SRN) with point supervision~\cite{dong2020scale} is also a kind of weakly supervised framework based on SRN structure.

Typical semi-supervised GANs are unable to function in the regression regime due to biases introduced when using a single prediction goal. DG-GAN~\cite{olmschenk2019dense} generalized semi-supervised generative adversarial network (GANs) from classification problems to regression for use in dense crowd counting, refer to Fig.~\ref{fig:semi}.
%refer to Fig.~\ref{fig:semiclass} and Fig.~\ref{fig:semicc}. 
This work allows the dual-goal GAN to benefit from unlabeled data in the training process. And~\cite{olmschenk2019generalizing} is an extension of DG-GAN, which proposed a novel loss function for feature contrasting and resulted in a discriminator that can distinguish between fake and real data based on feature statistics. However, weakly supervised crowd counting still requires annotations. Besides, it also requires task-specific knowledge to design effective neural networks and loss functions for leveraging weakly labeled data.

\begin{figure*}[t]
	\centering
	\includegraphics[width=1.0\textwidth]{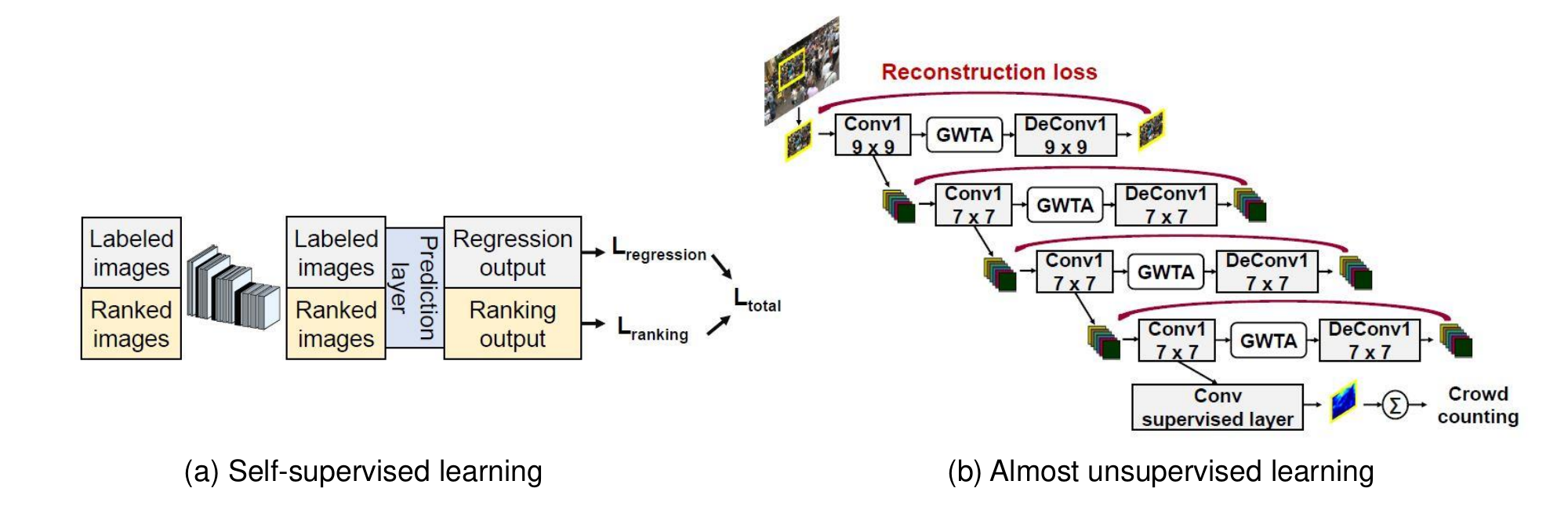}
	\caption{The workflow of self-supervised learning and almost unsupervised learning for crowd counting. (a) The architecture of L2R: a self-supervised learning setup~\cite{liu2018leveraging}. (b) The framework of GWTA-CCNN: an almost unsupervised learning method~\cite{sam2019almost}.}
	%~\cite{liu2018leveraging}.}
	\label{fig:unsupervised}
\end{figure*}

\subsection{Unsupervised and Self-supervised Learning}\label{data:unlabel}

%~\cite{sam2020completely}

%CNN-based 
Deep learning-based
approaches are highly data-driven, i.e., they require a large amount of diverse labeled data in the training process. The labeling process for crowd counting is expensive, but the unlabeled data are cheap and widely available~\cite{sam2020completely, liu2020towards}.
L2R~\cite{liu2018leveraging} leveraged abundantly available unlabeled crowd images in learning to rank framework, refer to %Fig.~\ref{fig:l2r},
Fig.~\ref{fig:unsupervised} (a),
which is based on the observation that any sub-image of a crowded scene image is guaranteed to contain the same number or fewer persons than the super-image. The pixel-wise regression loss is fused with the ranking regularization to learn better representation for crowd counting tasks on unlabeled data.

There is another potential direction to make use of unlabeled data such as the convolutional Winner-Take-All models, whose most parameters are obtained by unsupervised learning.
GWTA-CCNN~\cite{sam2019almost} utilized a 
Grid Winner-Take-All (GWTA) autoencoder to learn several layers of useful filters from unlabeled crowd images, refer to %Fig.~\ref{fig:almost}. 
Fig.~\ref{fig:unsupervised} (b). 
A small patch cropped from the original image is fed into the model. Most of the parameters are trained layer by layer based on the reconstruction loss. GWTA divides a convolution layer spatially into a grid of cells. Within each cell, only the maximumly activated neuron is allowed to update the filter. almost 99.9$\%$ of the parameters of the proposed model are trained without any labeled data, which the rest 0.1$\%$ are tuned with supervision. However, these kinds of self-supervised learning and almost unsupervised crowd counting approaches need a large amount of data to show effectiveness, which requires more training time and computational resources.

Lei et al.~\cite{lei2021towards} proposed a weakly supervised crowd counting method to train the model from a small number of dot-map annotations and a large number of count-level annotations, with is used to reducing the annotation cost for crowd counting. The key idea is to enforce the consistency between density maps and total object count on weakly labeled images as regularization terms.
The work of complete self-supervision ~\cite{sam2020completely} introduce a new training paradigm that does not need labeled data. This work reveals the power law nature for the distribution of crowds and adopt this signal for backpropagation in the optimal transport framework. This work achieves efficient crowd estimation.

\begin{table*}
	\centering
	%\doublespacing
	\caption{Comparisons of different supervisory signals for crowd counting. The representative schemes of different supervisory signals are analyzed based on their \textbf{advantages} and \textbf{limitations}.}
	\begin{tabular}{|c|c|c|c|}
		\hline
		\textbf{Category} &\textbf{Schemes}
		&\textbf{Advantages}  &\textbf{Limitations}
		\\
		\hline
		\multirow{2}*{\tabincell{c}{Fully Supervised \\learning}}
		&{CP-CNN~\cite{sindagi2017generating}}
		&{
		\tabincell{c}{Adaptive Gaussian kernel \\to accommodate \\different scales}
		}
		&{
		\tabincell{c}{Not flexible to \\non-rigid object}
		}
		\\
		\cline{2-4}
		~
		&{BL~\cite{ma2019bayesian}}
		&{\tabincell{c}{Bayesian loss \\to model \\non-rigid objects}}
		&{\tabincell{c}{More reliable supervision \\but suffers in \\large varied scales}}\\
		\hline
		\tabincell{c}{Weakly supervised and \\semi-supervised learning}
		&HA-CCN~\cite{sindagi2019ha}
		&\tabincell{c}{Low annotation \\complexity}
		&\tabincell{c}{Still requires weakly \\annotations and task \\specific knowledge}\\
		\hline
		\tabincell{c}{Unsupervised and \\self-supervised learning}
		&L2R~\cite{liu2018leveraging} 
		&\tabincell{c}{Low annotation cost; \\abundantly available}
        &\tabincell{c}{Large amount of \\data requires more \\training time}\\
		\hline
		\tabincell{c}{Automatic labeling \\through synthetic data}
		&CCWld~\cite{wang2019learning}
		&\tabincell{c}{Reduce labeling effort; \\enahnce accuracy; \\improve robustness}
		&\tabincell{c}{Large domain gap \\from synthetic \\to real data}\\
		\hline
	\end{tabular}	
	\label{tab:data-ana}
\end{table*}

\subsection{Automatic Labeling through Synthetic Data}\label{data:aug}

There are more challenges for crowd counting in the wild due to the changeable environment, large-range number of people cause the current methods can not work well. 
Due to scarce data, many methods suffer from over-fitting to a different extent.
Some researchers attempt to tackle this problem through synthetic data~\cite{hou2022enhancing,wang2021pixel}.
CCWld~\cite{wang2019learning} built a large-scale, diverse synthetic dataset, pretrain a crowd counter on the synthetic data, finetune on real data, propose a counting method via domain adaptation based cycle GAN, free humans from heavy data annotations.
The authors in~\cite{han2020focus} based on the GCC dataset, designed a better domain adaptation scheme for reducing the counting noise in the background area.
This paper pays more attention to the semantic consistency of the crowd and then could narrow the gap
using a large-scale human detection dataset to train a crowd, semantic model.
This method reduces the labeling effort, enhances accuracy, and improves robustness by making use of synthetic data. However, the synthetic data are still witnessed a larger domain gap compared with real data.

%\subsection{Domain Adaptation Crowd Counting}\label{data:adap}
\subsection{Domain Adaptive Crowd Counting}\label{data:adap}

Most of the existing crowd counting methods are designed in a specific domain. Thus, designing crowd counting models that can achieve high counting performance in any domain is a challenging but meaningful problem. There is some robust crowd counting approaches against domain shifts proposed in recent years~\cite{zhang2021cross, he2021error,  liu2022leveraging, yang2021class, zheng2021learning, reddy2020few}.

CVCS~\cite{zhang2021cross} proposes a cross-view cross-scene multi-view crowd counting paradigm, where the training and test set are from different scenes with arbitrary camera locations. CVCS are able to attentively selects and fuses multiple views using camera layout geometry, and a noise view regularization method to handle non-correspondence errors.
CDCC~\cite{wang2021neuron} proposes a neural linear transformation method, which exploits domain factor and biases weights to learn the domain shift.
AdaCrowd~\cite{reddy2021adacrowd} makes use of a crowd counting network and a guiding network, which predicts some parameters in the counting network based on the unlabeled data from a particular scene and adapt to the new scene.

The work of~\cite{gao2020feature} introduces a domain-adaptation-style crowd counting method by using multilevel feature-aware adaptation and structured density map alignment module, which is trained on generated data with ground-truth to the specific real-world scenes.
The work~\cite{gao2021domain} proposes to learn from synthetic crowd data and transferring knowledge to real data without ground truth. This DACC frame work adopt a high-quality image translation and density map reconstruction to enhance cross domain crowd counting quality. The work~\cite{cai2021leveraging} propose a two-step approach that captures the intra-domain knowledge to facilitate unsupervised cross-domain crowd counting via synthetic datasets.

The scale or density gap among datasets is another type of domain gap for domain adaptive crowd counting~\cite{ma2021towards, yan2021towards, wu2021dynamic, gong2022bi}. For example, 
The work of~\cite{ma2021towards} proposes a universal crowd counting model that can be applied across scenes and datasets via a scale alignment module. DCANet~\cite{yan2021towards} introduces a domain-guided channel attention network to guide the extraction of domain-specific feature representation for multi-domain crowd counting.
DKPNet~\cite{chen2021variational} designs a domain-specific knowledge propagating network for extracking knowledge from multiple domains for improving crowd counting performance.

\begin{table*}
	\centering
	\caption{Quantitative comparisons of state-of-the-art  crowd counting approaches with different supervisory signals. \textbf{Column} shows the type and number of columns for counting model. \textbf{Multi} is the Multi-column network; \textbf{Double} represents two columns; \textbf{Single} is the single column network.  \textbf{ST PartA} and \textbf{ST PartB} denotes ShanghaiTech A \& B dataset~\cite{zhang2016single}, respectively. The evaluation metrics for counting accuracy is \textbf{MAE} and \textbf{MSE}.}
	\begin{tabular}{|c|c|c|c|c|c|c|c|c|c|}
        \hline
        \multicolumn{2}{|c|}{\textbf{Typical Schemes}} 
        &\multicolumn{2}{|c|}{\textbf{ST PartA}} 
        &\multicolumn{2}{|c|}{\textbf{ST PartB}} 
        &\multicolumn{2}{|c|}{\textbf{UCF\_CC\_50}} 
		&\multicolumn{2}{|c|}{\textbf{UCF-QNRF}} \\
		\hline
		\textbf{Methods} %&\textbf{Year} 
		&\textbf{Column} &\textbf{MAE}  &\textbf{MSE} 
		&\textbf{MAE}  &\textbf{MSE} &\textbf{MAE}  &\textbf{MSE} 
		&\textbf{MAE}  &\textbf{MSE}\\
		\hline
		{CP-CNN~\cite{sindagi2017generating}} %&{2017} 
		&{Multi} 
		&{73.6} &{106.4} &{20.1} &{30.1}
		&{295.8} &{320.9} &{-} &{-}
		\\
		\hline
		\tabincell{c}{L2R~\cite{liu2018leveraging} (Query by example)} %&2018 
		&Double
        &72.0 &106.6 &14.4 &23.8
        &291.5 &397.6 
        & - & - \\
        \tabincell{c}{L2R~\cite{liu2018leveraging} (Query by keyword)} %&2018 
        &Double
        &73.6 &112.0 &13.7 & 21.4
        &279.6 &388.9 
        & - & - \\
        \hline
		BL~\cite{ma2019bayesian} %&2019 
		&Single
		&62.8 &101.8 &7.7 &12.7
		&229.3 &308.2 
		&88.7 &154.8 \\
		CCWld~\cite{wang2019learning} %&2019 
		&Single
		&64.8 &107.5 &7.6 &13.0 
		& - & - 
		&102.0 &171.4 \\
        URC~\cite{xu2021crowd} %&2021 
        &Single 
        & 72.8 & 111.6 & 12.0 & 18.7
        & 294.0 & 443.1 
        & 128.1 & 218.1 \\
        HA-CCN~\cite{sindagi2019ha} %&2020 
        &Single
        &58.3 &95.0 &6.7 &10.7
        &256.2 &348.4 
        &118.1 &180.4 \\
		\hline
	\end{tabular}	
	\label{tab:data-acc}
\end{table*}

\subsection{Comparisons}\label{data:summary}

We summarize different supervisory signals for crowd counting with their representative schemes
in Table~\ref{tab:data-ana}. We compare them in Table~\ref{tab:data-acc}.

BL achieves better performance on four different crowd counting datasets compared with CP-CNN, with a similar number of parameters for the backbone. The good performance of BL may be due to the Bayesian loss used to better model the non-rigid objects (e.g., people). The adaptive Gaussian kernel is widely used in crowd counting approaches, while the experimental results demonstrate the effectiveness of Bayesian loss, which is more reliable supervision.

CCWld shows much better accuracy than MCNN in Table~\ref{tab:data-acc} on various datasets with different backgrounds. We observe CCWld enhances the performance of counting accuracy and also improves the robustness, which is suitable for many real-world applications with diverse scenes, different view angles, and lighting conditions.

As shown in Table~\ref{tab:data-acc}, the performance of HA-CNN is much better than other state-of-the-arts. After carefully designing the deep neural networks and loss functions, weakly supervised crowd counting achieves much better accuracy with relatively low annotation complexity.

The MAE and MSE of L2R (query by example) and L2R (query by keyword) is lower than CP-CNN. This confirms that leveraging the abundantly available unlabeled data improves counting performance. The experimental results further demonstrate that making use of unlabeled data is a promising direction for crowd counting.

\subsection{Others}\label{data:others}
There are some other learning paradigms for crowd counting.

There is a typical training paradigm that is count from scalar representation. Some recent works achieve excellent results compared with density map regression method or learning from point map representation. TransCrowd~\cite{liang2022transcrowd} proposes to formulate crowd counting as a sequence-to-count paradigm based on transformers and achieves satisfactory performance.
CrowdMLP~\cite{wang2022crowdmlp} presents a multi-granularity MLP regressor for capturing global information and enchance crowd counting quality.

Recent research shows that the crowd localization can enhance the counting performance.
FIDT~\cite{liang2021focal} introduces a focal inverse distance transform map for crowd counting and crowd localization, which simultaneously conduct counting and crowd localization based on the FIDT map. IIM~\cite{gao2020learning} presents an independent instance map segmentation for crowd localization by segmenting people crowds into non-overlapped independent components.

There is another series of counting works that achieve crowd counting from remote sensing data. The work~\cite{zhu2021graph} introduces a crowd counting benchmark from remote sensing perspective. The work~\cite{gao2020counting} proposes a large-scale dense objects counting dataset based on remote sensing images. The work~\cite{zhao2020flow} proposes a flow-based Bi-path Network for remote sensing video sequences. IS-Count~\cite{meng2021count} presents a convariate-based importance sampling method for counting from remote sensing images. Compared with counting from normal perspective, the remote sensing images suffers more from small object recognition issues in designing the counting networks but the problem of scale variation for counting from normal perspective is more serious.

%------------------Conclusion---------------------------------------------
%\input{chap/conc}
\section{Conclusion and Future Directions}\label{sec:conc}

Crowd counting is an important and challenging problem in computer vision. This survey paper covers the design considerations and recent advances with respect to single image crowd counting problem, and summarizes more than 200 crowd counting schemes using deep learning approaches proposed since 2015.
We have discussed the major datasets, performance metrics, design considerations, techniques, and representative schemes to tackle the problem.
We provide a comprehensive overview and comparison of three major design modules for deep learning %models 
in crowd counting, deep neural network design, loss function, and supervisory signal.
The research field of crowd counting is rich and still evolving.
We discuss some future trends and possible research directions below:

\begin{itemize}
\item \textit{Automatic and lightweight network designing} has drawn much attention in recent years~\cite{liu2020efficient, shi2020real, wang2020mobilecount, wu2020fast}. Currently, designing CNN-based crowd counting models still requires a manual network and feature selection with strong domain knowledge. Automated Machine Learning %(AutoML) 
has been applied to image classification and object detection, which has the potential to automatically design efficient crowd counting architectures. 
Besides, CNN-based crowd counting models have increased in-depth with millions of parameters, which requires massive computation. Thus, there is also a need for model compression and acceleration techniques to deploy lightweight model.

\item \textit{Weakly supervised and unsupervised crowd counting} is able to reduce the labeling effort.
With the performance saturation for some supervised learning scenarios, researchers devote efforts to make use of unlabeled and weakly labeled images for crowd counting
Most of the state-of-the-art algorithms are based on fully supervised learning and trained with point-wise annotations, which has several limitations such as labor-intensive labeling process, easily over-fitting, and not salable in the absence of densely labeled crowd images. Weakly-supervised and unsupervised learning has attracted much attention in vision applications, which has value for crowd counting tasks to reduce labeling effort, enhance counting accuracy and improve robustness.

\item \textit{Crowd counting in videos} is becoming an active research direction. A straightforward approach is to consider the video frames independently by making use of the crowd counting techniques proposed for still images. This is not satisfactory because it ignores the continuity or temporal correlation between frames, i.e., the motion information. 
Bidirectional ConvLSTM~\cite{xiong2017spatiotemporal} is a recent attempt to leverage spatial-temporal information in video. 
There are some recent attempts to exploit the correlation in video data
~\cite{zou2019enhanced, fang2020multi, han2022dr, liu2020counting, ren2020tracking, ma2021spatiotemporal, liu2020estimating, gao2020feature}.
However, LSTM-based framework is not easy to train or to be extended to a general scenario. The 3D kernel is not effective in extracting the long-range contextual information.
Effectively making use of the temporal correlation for accurate and efficient near real-time crowd counting systems is also a potential research direction.

\item \textit{Multi-view fusion for crowd counting} is important as a single camera cannot capture large and wide areas (e.g., parks, public squares). Multiple cameras with overlapping view are required to solve the wide-area counting task. There are some recent multi-view fusion approaches for crowd counting~\cite{zhang2019wide}, which proposes a multi-camera fusion method to predict a ground-plane density map of the 3D world. There is also another approach based on a 2D-to-3D projection with 3D density map estimation and a 3D-to-2D projection consistency measure method~\cite{zhang20203d}.
Multi-view fusion for crowd counting provides a vivid visualization for the scenes, as well as the potentials for other applications like observing the scene in arbitrary view angles, which may contribute to better scene understanding. 
Therefore, crowd counting with multi-view fusion represents important research value.

\end{itemize}

\clearpage

\begin{table*}[t]
	\centering
	%\doublespacing
	\caption{A comprehensive performance analysis of various categories of crowd counting methods across different datasets. {\color{red}Red} denotes the best performance and {\color{blue}blue} denotes the third best performance. \textbf{ST PartA} is the ShanghaiTech A dataset~\cite{zhang2016single}. The evaluation metrics for the counting performance is \textbf{MAE} and \textbf{MSE}. 
	%FSL/PAL/SSL; Transcrowd (weak-supervised learning)
	}
	\begin{tabular}{|c|c|c|c|c|c|c|c|c|c|c|c|c|}
		\hline
		\multicolumn{3}{|c|}{\textbf{Typical Schemes}} & \multicolumn{2}{|c|}{\textbf{ST PartA}} & %\multicolumn{2}{|c|}{\textbf{ST PartB}} &
		\multicolumn{2}{|c|}{\textbf{UCF\_CC\_50}} &
		\multicolumn{2}{|c|}{\textbf{UCF-QNRF}} &
		\multicolumn{2}{|c|}{\textbf{NWPU}} 		
		\\
		\hline
		\textbf{\tabincell{c}{Methods}} &\textbf{Year} &\textbf{Column} &\textbf{MAE}  &\textbf{MSE} 
		&\textbf{MAE} &\textbf{MSE}
		&\textbf{MAE}  &\textbf{MSE} &\textbf{MAE}  &\textbf{MSE}  %&\textbf{MAE}  &\textbf{MSE}
		\\
		\hline
		\hline
		CSRNet~\cite{li2018csrnet} &2018 &Single &68.2 &115.0 %&10.6 &16.0 
		&266.1 &397.5 & - & - &104.8 &433.4 \\ 		SaCNN~\cite{zhang2018crowd} &2018 &Single &86.8 &139.2 %&16.2 &25.8 
		&314.9 &424.8 & - & - & - & - \\
		DADNet~\cite{guo2019dadnet} &2019 &Single &64.2 &99.9 %&8.8 &13.5 
		&285.5 &389.7 &113.2 &189.4 & - & - \\
		MRNet~\cite{tan2019crowd} &2019 &Single &63.3 &97.8 %&7.5 &11.5 
		&232.3 &314.8 &111.1 &182.8 & - & - \\		ADCNet~\cite{liu2019adcrowdnet} &2019 &Single &70.9 &115.2 %&7.7 &12.9 
		&273.6 &362.0 & - & - & - & - \\
		HA-CNN~\cite{sindagi2019ha} &2019 &Single &62.9 &94.9 %&8.1 &13.4 
		&256.2 &348.4 &118.1 &180.4 & - & - \\
		PGCNet~\cite{yan2019perspective} &2019 &Single &57.0 &{\color{blue}86.0} %&8.8 &13.7 
		&244.6 &361.2 & - & - & - & - \\
		SDANet~\cite{miao2020shallow} &2020 &Single &63.6 &101.8 %&7.8 &10.2 
		&227.6 &316.4 & - & - & - & -  \\
		CTN~\cite{ranjan2020uncertainty} &2020 &Single &61.5 &103.4 %&7.5 &11.9 
		&210.0 &305.4 &86.0 &146.0 &78.0 &448.0 \\	
		DM-Count~\cite{wang2020distribution} &2020 &Single &59.7 &95.7 %&7.4 &11.8 
		&211.0 &291.5  &85.6 &148.3   &70.5 &357.6   \\		
		NAS-Count~\cite{hu2020count} &2020 &Single &56.7 &93.4 %&6.7 &10.2 
		&208.4 &297.3 &101.8 &163.2 & - & -  \\	
		SRF-Net~\cite{chen2020scale} &2020 &Single &60.4 &97.2 %&7.1 &11.5 
		&197.3 &271.8 &98.0 &170.0 & - & -   \\	
		ADSCNet~\cite{bai2020adaptive} &2020 &Single &55.4 &97.7 %&6.4 &11.3 
		&198.4 &267.3 &{\color{red}71.3} &132.5 & - & -  \\		
		UEPNet~\cite{wang2021uniformity} &2021 &Single &54.6 &91.2 %&6.4 &10.9 
		&165.2 &275.9 &81.1 &131.7 & - & -   \\
		S3~\cite{lin2021direct} &2021 &Single &57.0 &96.0 %&{\color{blue}6.3} &10.6 
		& - & - &80.6 &139.8 &83.5 &346.9 \\
		NDConv~\cite{zand2022multiscale} &2022 &Single &61.4 &104.2 %&7.8 &13.8 
		&167.2 &240.6 &95.9 &182.4 & - & - \\	TransCrowd~\cite{liang2022transcrowd} &2022 &Single &66.1 &105.1 %&9.3 &16.1 
		&272.2 &395.3 &97.2 &168.5 &88.4 &400.5 \\	
		MAN~\cite{lin2022boosting} &2022 &Single &56.8 &90.3 
		%& - & - 
		& - & - &77.3 &131.5 &76.5 &{\color{blue}323.0}
	
		\\
		\hline
		CMTL~\cite{sindagi2017cnn} &2017 &Double &101.3 &152.4 %&20.0 &31.1 
		&322.8 &397.9 & - & - & - & - \\
		ACSCP~\cite{shen2018crowd} &2018 &Double &75.7 &102.7 %&17.2 &27.4 
		&291.0 &404.6 & - & - & - & - \\		
		SDNet~\cite{ma2021towards} &2021 &Double &55.0 &92.7 %& - & - 
		&197.5 &264.1 &80.7 &146.3 & - & - \\
		BM-Count~\cite{liu2021bipartite} &2021 &Double &57.3 &90.7 %&7.3 &11.4 
		& - & - &81.2 &138.6 &83.4 &358.4 \\
		BSCC~\cite{modolo2021understanding} &2021 &Double &58.3 &100.1 %&6.7 &10.7 
		& - & - &86.3 &153.1 & - & -  \\
		P2PNet~\cite{song2021rethinking} &2021 &Double &{\color{red}52.7} &{\color{red}85.1} %&{\color{blue}6.3} &{\color{blue}9.9} 
		&172.7 &256.2 &85.3 &154.5 &77.4 &362.0 \\
		GauNet~\cite{cheng2022rethinking} &2022 &Double &54.8 &89.1 %&{\color{red}6.2} &{\color{blue}9.9} 
		&186.3 &256.5 &81.6 &153.7 & - & - \\	
		RAN~\cite{chen2022region} &2022 &Double &57.9 &99.2 %&7.2 &11.9 
		&{\color{red}155.0} &{\color{red}219.5} &83.4 &141.8 &{\color{blue}65.3} &432.9 \\	
		\hline
		MCNN~\cite{zhang2016single} &2016 &Multi &110.2 &173.2 %&26.4 &41.3 
		&377.6 &509.1 &277 &426 &218.5 &700.6 \\
		CP-CNN~\cite{sindagi2017generating} &2017 &Multi &73.6 &106.4 %&20.1 &30.1 
		&295.8 &320.9 & - & - & - & -   \\
		Switching~\cite{babu2017switching} &2017 &Multi &90.4 &135.0 %&21.6 &33.4 
		&318.1 &439.2 & - & - & - & - \\
		SANet~\cite{cao2018scale} &2018 &Multi &67.0 &104.5 %&8.4 &13.6 
		&258.4 &334.9 & - & - & - & - \\
		DSSINet~\cite{liu2019crowd} &2019 &Multi &60.6 &96.0 %&6.9 &10.3 
		& 216.9 &302.4 &99.1 &159.2 & - & - \\
		CFF~\cite{shi2019counting} &2019 &Multi &65.2 &109.4 %&7.2 &12.2 
		& - & - &93.8 &146.5 & - & - \\
		S-DCNet~\cite{xiong2019open} &2019 &Multi &58.3 &95.0 %&6.7 &10.7 
		&204.2 &301.3 &104.4 &176.1 & - & - \\
		CAN~\cite{liu2019context} &2019 &Multi &62.3 &100.0 %&7.8 &12.2 
		&212.2 &243.7 &107.0 &183.0 & - & - \\
		SPANet~\cite{cheng2019learning} &2019 &Multi &59.4 &92.5 %&6.5 &9.9 
		&232.6 &311.7 & - & - & - & - \\		
		DPN~\cite{ma2020learning} &2020 &Multi &58.1 &91.7 %&6.5 &10.1 
		&183.2 &284.5 &84.7 &147.2  & - & -   \\
		AMRNet~\cite{liu2020adaptive} &2020 &Multi &61.6 &98.4 %&7.0 &11.0 
		&184.0 &265.8 &86.6 &152.2 & - & - \\
		ASNet~\cite{jiang2020attention} &2020 &Multi &57.8 &90.1 %& - & - 
		&174.8 &251.6 &91.6 &159.7 & - & - \\
		DeepCount~\cite{chen2019deep} &2020 &Multi &65.2 &112.5 %&7.2 &11.3 
		& - & - &95.7 &167.1 & - & - \\
		ikNN~\cite{olmschenk2019improving} &2020 &Multi &68.0 &117.7 %&13.4 &21.4 
		&237.8 &305.7 &104.0  &172.0 & - & -  \\
		M-SFANet~\cite{thanasutives2021encoder} &2020 &Multi &57.6 &94.5 %&{\color{blue}6.3} &10.1 
		&167.5 &256.3 &87.6 &147.8 & - & - \\
		EPA~\cite{yang2020embedding} &2021 &Multi &60.9 &91.6 %&7.9 &11.6 
		&250.1 &352.1 & - & - & - & - \\
		DKPNet~\cite{chen2021variational} & 2021 &Multi &55.6 &91.0 %&6.6 &10.9 
		& - & - &81.4 &147.2 &{\color{red}61.8} &438.7 \\	
		SASNet~\cite{song2021choose} &2021 &Multi &{\color{blue}53.6} &88.4 %&6.4 &{\color{blue}9.9} 
		&{\color{blue}161.4} &{\color{blue}234.5} &85.2 &147.3 & - & - \\
		MFDC~\cite{liu2021exploiting} &2021 &Multi &55.4 &91.3 %&6.9 &10.3 
		& - & - &{\color{blue}76.2} &{\color{blue}121.5} &74.7 &{\color{red}267.9} \\		
		MPS~\cite{zand2022multiscale} &2022 &Multi &71.4 &110.7 %&9.6 &15.0 
		& - & - & - & - & - & - \\
		\hline
		MNA~\cite{wan2020modeling} &2020 &N/A &61.9 &99.6 
		%&7.4 &11.3 
		& - & - & 85.8 &150.6 &96.9 &534.2  \\
		BL~\cite{ma2019bayesian} &2019 &N/A &62.8 &101.8 
		%&7.7 &12.7 
		&229.3 &308.2 & 88.7 &154.8 & - & -  \\		
		UOT~\cite{ma2021learning} &2021 &N/A &58.1 &95.9 
		%&6.5 &10.2 
		& - & - &83.3 &142.3 &87.8 &387.5 \\
		BinLoss~\cite{shivapuja2021wisdom} &2021 &N/A &61.3 &88.7 
		%&7.4 &{\color{red}9.2} 
		& - & - &85.9 &{\color{red}120.6} &71.7 &376.4  \\
		\hline

		\hline
		%\hline
	\end{tabular}	
	\label{tab:quan}
\end{table*}

\clearpage

\bibliographystyle{spmpsci}

\bibliography{survey}

\begin{thebibliography}{100}
\providecommand{\url}[1]{{#1}}
\providecommand{\urlprefix}{URL }
\expandafter\ifx\csname urlstyle\endcsname\relax
  \providecommand{\doi}[1]{DOI~\discretionary{}{}{}#1}\else
  \providecommand{\doi}{DOI~\discretionary{}{}{}\begingroup
  \urlstyle{rm}\Url}\fi

\bibitem{abousamra2021localization}
Abousamra, S., Hoai, M., Samaras, D., Chen, C.: Localization in the crowd with
  topological constraints.
\newblock In: AAAI (2021)

\bibitem{ahuja2019survey}
Ahuja, K.R., Charniya, N.N.: A survey of recent advances in crowd density
  estimation using image processing.
\newblock In: ICCES (2019)

\bibitem{amirgholipour2020pdanet}
Amirgholipour, S., He, X., Jia, W., Wang, D., Liu, L.: Pdanet: Pyramid
  density-aware attention net for accurate crowd counting.
\newblock NeuroComputing  (2020)

\bibitem{arteta2016counting}
Arteta, C., Lempitsky, V., Zisserman, A.: Counting in the wild.
\newblock In: ECCV (2016)

\bibitem{aydin2019deep}
Ayd{\i}n, S.: Deep learning classification of neuro-emotional phase domain
  complexity levels induced by affective video film clips.
\newblock IEEE Journal of Biomedical and Health Informatics  (2019)

\bibitem{babu2017switching}
Babu~Sam, D., Surya, S., Venkatesh~Babu, R.: Switching convolutional neural
  network for crowd counting.
\newblock In: CVPR (2017)

\bibitem{bai2019crowd}
Bai, H., Wen, S., Gary~Chan, S.H.: Crowd counting on images with scale
  variation and isolated clusters.
\newblock In: ICCV Workshops (2019)

\bibitem{bai2020adaptive}
Bai, S., He, Z., Qiao, Y., Hu, H., Wu, W., Yan, J.: Adaptive dilated network
  with self-correction supervision for counting.
\newblock In: CVPR (2020)

\bibitem{von2016gaussian}
von Borstel, M., Kandemir, M., Schmidt, P., Rao, M.K., Rajamani, K., Hamprecht,
  F.A.: Gaussian process density counting from weak supervision.
\newblock In: ECCV (2016)

\bibitem{cai2021leveraging}
Cai, Y., Chen, L., Ma, Z., Lu, C., Wang, C., He, G.: Leveraging intra-domain
  knowledge to strengthen cross-domain crowd counting.
\newblock In: ICME (2021)

\bibitem{cao2018scale}
Cao, X., Wang, Z., Zhao, Y., Su, F.: Scale aggregation network for accurate and
  efficient crowd counting.
\newblock In: ECCV (2018)

\bibitem{chan2008privacy}
Chan, A.B., Liang, Z.S.J., Vasconcelos, N.: Privacy preserving crowd
  monitoring: Counting people without people models or tracking.
\newblock In: CVPR (2008)

\bibitem{chan2009bayesian}
Chan, A.B., Vasconcelos, N.: Bayesian poisson regression for crowd counting.
\newblock In: ICCV (2009)

\bibitem{chan2012counting}
Chan, A.B., Vasconcelos, N.: Counting people with low-level features and
  bayesian regression.
\newblock TIP  (2012)

\bibitem{chen2021variational}
Chen, B., Yan, Z., Li, K., Li, P., Wang, B., Zuo, W., Zhang, L.: Variational
  attention: Propagating domain-specific knowledge for multi-domain learning in
  crowd counting.
\newblock In: ICCV (2021)

\bibitem{chen2020crowd}
Chen, J., Su, W., Wang, Z.: Crowd counting with crowd attention convolutional
  neural network.
\newblock Neurocomputing  (2020)

\bibitem{chen2012feature}
Chen, K., Loy, C.C., Gong, S., Xiang, T.: Feature mining for localised crowd
  counting.
\newblock In: BMVC (2012)

\bibitem{chen2020relevant}
Chen, X., Bin, Y., Gao, C., Sang, N., Tang, H.: Relevant region prediction for
  crowd counting.
\newblock Neurocomputing  (2020)

\bibitem{chen2019scale}
Chen, X., Bin, Y., Sang, N., Gao, C.: Scale pyramid network for crowd counting.
\newblock In: WACV (2019)

\bibitem{chen2020scale}
Chen, Y., Gao, C., Su, Z., He, X., Liu, N.: Scale-aware rolling fusion network
  for crowd counting.
\newblock In: ICME (2020)

\bibitem{chen2022region}
Chen, Y., Yang, J., Zhang, D., Zhang, K., Chen, B., Du, S.: Region-aware
  network: Model human’s top-down visual perception mechanism for crowd
  counting.
\newblock Neural Networks  (2022)

\bibitem{chen2019deep}
Chen, Z., Cheng, J., Yuan, Y., Liao, D., Li, Y., Lv, J.: Deep density-aware
  count regressor.
\newblock ECAI  (2019)

\bibitem{cheng2022rethinking}
Cheng, Z.Q., Dai, Q., Li, H., Song, J., Wu, X., Hauptmann, A.G.: Rethinking
  spatial invariance of convolutional networks for object counting.
\newblock In: CVPR (2022)

\bibitem{cheng2019learning}
Cheng, Z.Q., Li, J.X., Dai, Q., Wu, X., Hauptmann, A.G.: Learning spatial
  awareness to improve crowd counting.
\newblock In: ICCV (2019)

\bibitem{cheng2019improving}
Cheng, Z.Q., Li, J.X., Dai, Q., Wu, X., He, J.Y., Hauptmann, A.G.: Improving
  the learning of multi-column convolutional neural network for crowd counting.
\newblock In: ACM Multimedia (2019)

\bibitem{chrysler2021literature}
Chrysler, A., Gunarso, R., Puteri, T., Warnars, H.: A literature review of
  crowd-counting system on convolutional neural network.
\newblock In: IOP Conference Series: Earth and Environmental Science (2021)

\bibitem{dai2021dense}
Dai, F., Liu, H., Ma, Y., Zhang, X., Zhao, Q.: Dense scale network for crowd
  counting.
\newblock In: International Conference on Multimedia Retrieval (2021)

\bibitem{deb2018aggregated}
Deb, D., Ventura, J.: An aggregated multicolumn dilated convolution network for
  perspective-free counting.
\newblock In: CVPR Workshops (2018)

\bibitem{ding2020crowd}
Ding, X., He, F., Lin, Z., Wang, Y., Guo, H., Huang, Y.: Crowd density
  estimation using fusion of multi-layer features.
\newblock IEEE Transactions on Intelligent Transportation Systems  (2020)

\bibitem{ding2018deeply}
Ding, X., Lin, Z., He, F., Wang, Y., Huang, Y.: A deeply-recursive
  convolutional network for crowd counting.
\newblock In: ICASSP (2018)

\bibitem{dong2020crowd}
Dong, L., Zhang, H., Ji, Y., Ding, Y.: Crowd counting by using multi-level
  density-based spatial information: A multi-scale cnn framework.
\newblock Information Sciences  (2020)

\bibitem{dong2020scale}
Dong, Z., Zhang, R., Shao, X., Li, Y.: Scale-recursive network with point
  supervision for crowd scene analysis.
\newblock Neurocomputing  (2020)

\bibitem{draghici2018survey}
Draghici, A., Steen, M.V.: A survey of techniques for automatically sensing the
  behavior of a crowd.
\newblock ACM Computing Surveys  (2018)

\bibitem{du2020visdrone}
Du, D., Wen, L., Zhu, P., Fan, H., Hu, Q., Ling, H., Shah, M., Pan, J., Al-Ali,
  A., Mohamed, A., et~al.: Visdrone-cc2020: The vision meets drone crowd
  counting challenge results.
\newblock In: ECCV (2020)

\bibitem{duan2020sofa}
Duan, H., Wang, S., Guan, Y.: Sofa-net: Second-order and first-order attention
  network for crowd counting.
\newblock BMVC  (2020)

\bibitem{fang2020multi}
Fang, Y., Gao, S., Li, J., Luo, W., He, L., Hu, B.: Multi-level feature fusion
  based locality-constrained spatial transformer network for video crowd
  counting.
\newblock Neurocomputing  (2020)

\bibitem{fang2019locality}
Fang, Y., Zhan, B., Cai, W., Gao, S., Hu, B.: Locality-constrained spatial
  transformer network for video crowd counting.
\newblock In: ICME (2019)

\bibitem{gao2020counting}
Gao, G., Liu, Q., Wang, Y.: Counting dense objects in remote sensing images.
\newblock In: ICASSP (2020)

\bibitem{gao2020learning}
Gao, J., Han, T., Yuan, Y., Wang, Q.: Learning independent instance maps for
  crowd localization.
\newblock arXiv preprint arXiv:2012.04164  (2020)

\bibitem{gao2021domain}
Gao, J., Han, T., Yuan, Y., Wang, Q.: Domain-adaptive crowd counting via
  high-quality image translation and density reconstruction.
\newblock TNNLS  (2021)

\bibitem{gao2019pcc}
Gao, J., Wang, Q., Li, X.: Pcc net: Perspective crowd counting via spatial
  convolutional network.
\newblock TCSVT  (2019)

\bibitem{gao2019scar}
Gao, J., Wang, Q., Yuan, Y.: Scar: Spatial-/channel-wise attention regression
  networks for crowd counting.
\newblock Neurocomputing  (2019)

\bibitem{gao2020feature}
Gao, J., Yuan, Y., Wang, Q.: Feature-aware adaptation and density alignment for
  crowd counting in video surveillance.
\newblock IEEE transactions on cybernetics  (2020)

\bibitem{gong2022bi}
Gong, S., Zhang, S., Yang, J., Dai, D., Schiele, B.: Bi-level alignment for
  cross-domain crowd counting.
\newblock In: CVPR (2022)

\bibitem{guerrero2015extremely}
Guerrero-G{\'o}mez-Olmedo, R., Torre-Jim{\'e}nez, B., L{\'o}pez-Sastre, R.,
  Maldonado-Basc{\'o}n, S., Onoro-Rubio, D.: Extremely overlapping vehicle
  counting.
\newblock In: Iberian Conference on Pattern Recognition and Image Analysis
  (2015)

\bibitem{guo2019dadnet}
Guo, D., Li, K., Zha, Z.J., Wang, M.: Dadnet: Dilated-attention-deformable
  convnet for crowd counting.
\newblock In: ACM Multimedia (2019)

\bibitem{han2022dr}
Han, T., Bai, L., Gao, J., Wang, Q., Ouyang, W.: Dr. vic: Decomposition and
  reasoning for video individual counting.
\newblock In: CVPR (2022)

\bibitem{han2020focus}
Han, T., Gao, J., Yuan, Y., Wang, Q.: Focus on semantic consistency for
  cross-domain crowd understanding.
\newblock In: ICASSP (2020)

\bibitem{he2019dynamic}
He, G., Ma, Z., Huang, B., Sheng, B., Yuan, Y.: Dynamic region division for
  adaptive learning pedestrian counting.
\newblock In: ICME (2019)

\bibitem{he2016deep}
He, K., Zhang, X., Ren, S., Sun, J.: Deep residual learning for image
  recognition.
\newblock In: CVPR (2016)

\bibitem{he2021error}
He, Y., Ma, Z., Wei, X., Hong, X., Ke, W., Gong, Y.: Error-aware density
  isomorphism reconstruction for unsupervised cross-domain crowd counting.
\newblock In: AAAI (2021)

\bibitem{hossain2019crowd}
Hossain, M., Hosseinzadeh, M., Chanda, O., Wang, Y.: Crowd counting using
  scale-aware attention networks.
\newblock In: WACV (2019)

\bibitem{hou2022enhancing}
Hou, Y., Li, C., Lu, Y., Zhu, L., Li, Y., Jia, H., Xie, X.: Enhancing and
  dissecting crowd counting by synthetic data.
\newblock In: ICASSP (2022)

\bibitem{hou2020bba}
Hou, Y., Li, C., Yang, F., Ma, C., Zhu, L., Li, Y., Jia, H., Xie, X.: Bba-net:
  A bi-branch attention network for crowd counting.
\newblock In: ICASSP (2020)

\bibitem{hu2020count}
Hu, Y., Jiang, X., Liu, X., Zhang, B., Han, J., Cao, X., Doermann, D.:
  Nas-count: Counting-by-density with neural architecture search.
\newblock In: ECCV (2020)

\bibitem{huang2020stacked}
Huang, S., Li, X., Cheng, Z.Q., Zhang, Z., Hauptmann, A.: Stacked pooling for
  boosting scale invariance of crowd counting.
\newblock In: ICASSP (2020)

\bibitem{huang2017body}
Huang, S., Li, X., Zhang, Z., Wu, F., Gao, S., Ji, R., Han, J.: Body structure
  aware deep crowd counting.
\newblock TIP  (2017)

\bibitem{huberman2022single}
Huberman-Spiegelglas, I., Fattal, R.: Single image object counting and
  localizing using active-learning.
\newblock In: WACV (2022)

\bibitem{idrees2013multi}
Idrees, H., Saleemi, I., Seibert, C., Shah, M.: Multi-source multi-scale
  counting in extremely dense crowd images.
\newblock In: CVPR (2013)

\bibitem{idrees2018composition}
Idrees, H., Tayyab, M., Athrey, K., Zhang, D., Al-Maadeed, S., Rajpoot, N.,
  Shah, M.: Composition loss for counting, density map estimation and
  localization in dense crowds.
\newblock In: ECCV (2018)

\bibitem{jiang2019mask}
Jiang, S., Lu, X., Lei, Y., Liu, L.: Mask-aware networks for crowd counting.
\newblock TCSVT  (2019)

\bibitem{jiang2019crowd}
Jiang, X., Xiao, Z., Zhang, B., Zhen, X., Cao, X., Doermann, D., Shao, L.:
  Crowd counting and density estimation by trellis encoder-decoder networks.
\newblock In: CVPR (2019)

\bibitem{jiang2019learning}
Jiang, X., Zhang, L., Lv, P., Guo, Y., Zhu, R., Li, Y., Pang, Y., Li, X., Zhou,
  B., Xu, M.: Learning multi-level density maps for crowd counting.
\newblock TNNLS  (2019)

\bibitem{jiang2020attention}
Jiang, X., Zhang, L., Xu, M., Zhang, T., Lv, P., Zhou, B., Yang, X., Pang, Y.:
  Attention scaling for crowd counting.
\newblock In: CVPR (2020)

\bibitem{jiang2020density}
Jiang, X., Zhang, L., Zhang, T., Lv, P., Zhou, B., Pang, Y., Xu, M., Xu, C.:
  Density-aware multi-task learning for crowd counting.
\newblock IEEE Transactions on Multimedia  (2020)

\bibitem{kang2018crowd}
Kang, D., Chan, A.: Crowd counting by adaptively fusing predictions from an
  image pyramid.
\newblock BMVC  (2018)

\bibitem{khaki2021deepcorn}
Khaki, S., Pham, H., Han, Y., Kuhl, A., Kent, W., Wang, L.: Deepcorn: A
  semi-supervised deep learning method for high-throughput image-based corn
  kernel counting and yield estimation.
\newblock Knowledge-Based Systems  (2021)

\bibitem{kong2020weakly}
Kong, X., Zhao, M., Zhou, H., Zhang, C.: Weakly supervised crowd-wise attention
  for robust crowd counting.
\newblock In: ICASSP (2020)

\bibitem{krizhevsky2012imagenet}
Krizhevsky, A., Sutskever, I., Hinton, G.E.: Imagenet classification with deep
  convolutional neural networks.
\newblock NIPS  (2012)

\bibitem{kumar2019mtcnet}
Kumar, A., Jain, N., Tripathi, S., Singh, C., Krishna, K.: Mtcnet: Multi-task
  learning paradigm for crowd count estimation.
\newblock IEEE AVSS  (2019)

\bibitem{laradji2018blobs}
Laradji, I.H., Rostamzadeh, N., Pinheiro, P.O., Vazquez, D., Schmidt, M.: Where
  are the blobs: Counting by localization with point supervision.
\newblock In: ECCV (2018)

\bibitem{lei2021towards}
Lei, Y., Liu, Y., Zhang, P., Liu, L.: Towards using count-level weak
  supervision for crowd counting.
\newblock Pattern Recognition  (2021)

\bibitem{lempitsky2010learning}
Lempitsky, V., Zisserman, A.: Learning to count objects in images.
\newblock In: NIPS (2010)

\bibitem{li2021approaches}
Li, B., Huang, H., Zhang, A., Liu, P., Liu, C.: Approaches on crowd counting
  and density estimation: a review.
\newblock Pattern Analysis and Applications  (2021)

\bibitem{li2019cross}
Li, J., Xue, Y., Wang, W., Ouyang, G.: Cross-level parallel network for crowd
  counting.
\newblock IEEE Transactions on Industrial Informatics  (2019)

\bibitem{li2008estimating}
Li, M., Zhang, Z., Huang, K., Tan, T.: Estimating the number of people in
  crowded scenes by mid based foreground segmentation and head-shoulder
  detection.
\newblock In: ICPR (2008)

\bibitem{li2014crowded}
Li, T., Chang, H., Wang, M., Ni, B., Hong, R., Yan, S.: Crowded scene analysis:
  A survey.
\newblock TCSVT  (2014)

\bibitem{li2021learning}
Li, W., Cao, Z., Wang, Q., Chen, S., Feng, R.: Learning error-driven curriculum
  for crowd counting.
\newblock In: ICPR (2021)

\bibitem{li2019coda}
Li, W., Yongbo, L., Xiangyang, X.: Coda: Counting objects via scale-aware
  adversarial density adaption.
\newblock In: ICME (2019)

\bibitem{li2018csrnet}
Li, Y., Zhang, X., Chen, D.: Csrnet: Dilated convolutional neural networks for
  understanding the highly congested scenes.
\newblock In: CVPR (2018)

\bibitem{lian2021locating}
Lian, D., Chen, X., Li, J., Luo, W., Gao, S.: Locating and counting heads in
  crowds with a depth prior.
\newblock TPAMI  (2021)

\bibitem{liang2022transcrowd}
Liang, D., Chen, X., Xu, W., Zhou, Y., Bai, X.: Transcrowd: weakly-supervised
  crowd counting with transformers.
\newblock Science China Information Sciences  (2022)

\bibitem{liang2021focal}
Liang, D., Xu, W., Zhu, Y., Zhou, Y.: Focal inverse distance transform maps for
  crowd localization and counting in dense crowd.
\newblock arXiv preprint arXiv:2102.07925  (2021)

\bibitem{lin2021direct}
Lin, H., Hong, X., Ma, Z., Wei, X., Qiu, Y., Wang, Y., Gong, Y.: Direct measure
  matching for crowd counting.
\newblock IJCAI  (2021)

\bibitem{lin2022boosting}
Lin, H., Ma, Z., Ji, R., Wang, Y., Hong, X.: Boosting crowd counting via
  multifaceted attention.
\newblock In: CVPR (2022)

\bibitem{lin2010shape}
Lin, Z., Davis, L.S.: Shape-based human detection and segmentation via
  hierarchical part-template matching.
\newblock TPAMI  (2010)

\bibitem{ling2019indoor}
Ling, M., Geng, X.: Indoor crowd counting by mixture of gaussians label
  distribution learning.
\newblock TIP  (2019)

\bibitem{liu2019recurrent}
Liu, C., Weng, X., Mu, Y.: Recurrent attentive zooming for joint crowd counting
  and precise localization.
\newblock In: CVPR (2019)

\bibitem{liu2021bipartite}
Liu, H., Zhao, Q., Ma, Y., Dai, F.: Bipartite matching for crowd counting with
  point supervision.
\newblock In: IJCAI (2021)

\bibitem{liu2018decidenet}
Liu, J., Gao, C., Meng, D., Hauptmann, A.G.: Decidenet: Counting varying
  density crowds through attention guided detection and density estimation.
\newblock In: CVPR (2018)

\bibitem{liu2020efficient}
Liu, L., Chen, J., Wu, H., Chen, T., Li, G., Lin, L.: Efficient crowd counting
  via structured knowledge transfer.
\newblock In: ACM Multimedia (2020)

\bibitem{liu2021cross}
Liu, L., Chen, J., Wu, H., Li, G., Li, C., Lin, L.: Cross-modal collaborative
  representation learning and a large-scale rgbt benchmark for crowd counting.
\newblock In: CVPR (2021)

\bibitem{liu2020denet}
Liu, L., Jia, W., Jiang, J., Amirgholipour, S., Wang, Y., Zeibots, M., He, X.:
  Denet: A universal network for counting crowd with varying densities and
  scales.
\newblock IEEE Transactions on Multimedia  (2020)

\bibitem{liu2020weighing}
Liu, L., Lu, H., Zou, H., Xiong, H., Cao, Z., Shen, C.: Weighing counts:
  Sequential crowd counting by reinforcement learning.
\newblock In: ECCV (2020)

\bibitem{liu2019crowd}
Liu, L., Qiu, Z., Li, G., Liu, S., Ouyang, W., Lin, L.: Crowd counting with
  deep structured scale integration network.
\newblock In: ICCV (2019)

\bibitem{liu2018crowd}
Liu, L., Wang, H., Li, G., Ouyang, W., Lin, L.: Crowd counting using deep
  recurrent spatial-aware network.
\newblock IJCAI  (2018)

\bibitem{liu2019adcrowdnet}
Liu, N., Long, Y., Zou, C., Niu, Q., Pan, L., Wu, H.: Adcrowdnet: An
  attention-injective deformable convolutional network for crowd understanding.
\newblock In: CVPR (2019)

\bibitem{liu2022leveraging}
Liu, W., Durasov, N., Fua, P.: Leveraging self-supervision for cross-domain
  crowd counting.
\newblock In: CVPR (2022)

\bibitem{liu2019context}
Liu, W., Salzmann, M., Fua, P.: Context-aware crowd counting.
\newblock In: CVPR (2019)

\bibitem{liu2020counting}
Liu, W., Salzmann, M., Fua, P.: Counting people by estimating people flows.
\newblock TPAMI  (2020)

\bibitem{liu2020estimating}
Liu, W., Salzmann, M., Fua, P.: Estimating people flows to better count them in
  crowded scenes.
\newblock In: ECCV (2020)

\bibitem{liu2021exploiting}
Liu, X., Li, G., Han, Z., Zhang, W., Yang, Y., Huang, Q., Sebe, N.: Exploiting
  sample correlation for crowd counting with multi-expert network.
\newblock In: ICCV (2021)

\bibitem{liu2018leveraging}
Liu, X., Van De~Weijer, J., Bagdanov, A.D.: Leveraging unlabeled data for crowd
  counting by learning to rank.
\newblock In: CVPR (2018)

\bibitem{liu2020adaptive}
Liu, X., Yang, J., Ding, W., Wang, T., Wang, Z., Xiong, J.: Adaptive mixture
  regression network with local counting map for crowd counting.
\newblock In: ECCV (2020)

\bibitem{liu2020semi}
Liu, Y., Liu, L., Wang, P., Zhang, P., Lei, Y.: Semi-supervised crowd counting
  via self-training on surrogate tasks.
\newblock In: ECCV (2020)

\bibitem{liu2019point}
Liu, Y., Shi, M., Zhao, Q., Wang, X.: Point in, box out: Beyond counting
  persons in crowds.
\newblock In: CVPR (2019)

\bibitem{liu2020towards}
Liu, Y., Wang, Z., Shi, M., Satoh, S., Zhao, Q., Yang, H.: Towards unsupervised
  crowd counting via regression-detection bi-knowledge transfer.
\newblock In: ACM Multimedia (2020)

\bibitem{liu2020crowd}
Liu, Y., Wen, Q., Chen, H., Liu, W., Qin, J., Han, G., He, S.: Crowd counting
  via cross-stage refinement networks.
\newblock IEEE Transactions on Image Processing  (2020)

\bibitem{louedec2021gaussian}
Lou{\"e}dec, J.L., Cielniak, G.: Gaussian map predictions for 3d surface
  feature localisation and counting.
\newblock BMVC  (2021)

\bibitem{luo2020hybrid}
Luo, A., Yang, F., Li, X., Nie, D., Jiao, Z., Zhou, S., Cheng, H.: Hybrid graph
  neural networks for crowd counting.
\newblock In: AAAI (2020)

\bibitem{ma2019atrous}
Ma, J., Dai, Y., Tan, Y.P.: Atrous convolutions spatial pyramid network for
  crowd counting and density estimation.
\newblock Neurocomputing  (2019)

\bibitem{ma2021spatiotemporal}
Ma, Y.J., Shuai, H.H., Cheng, W.H.: Spatiotemporal dilated convolution with
  uncertain matching for video-based crowd estimation.
\newblock IEEE Transactions on Multimedia  (2021)

\bibitem{ma2021towards}
Ma, Z., Hong, X., Wei, X., Qiu, Y., Gong, Y.: Towards a universal model for
  cross-dataset crowd counting.
\newblock In: ICCV (2021)

\bibitem{ma2019bayesian}
Ma, Z., Wei, X., Hong, X., Gong, Y.: Bayesian loss for crowd count estimation
  with point supervision.
\newblock In: ICCV (2019)

\bibitem{ma2020learning}
Ma, Z., Wei, X., Hong, X., Gong, Y.: Learning scales from points: A scale-aware
  probabilistic model for crowd counting.
\newblock In: ACM Multimedia (2020)

\bibitem{ma2021learning}
Ma, Z., Wei, X., Hong, X., Lin, H., Qiu, Y., Gong, Y.: Learning to count via
  unbalanced optimal transport.
\newblock In: AAAI (2021)

\bibitem{mao2021pyramid}
Mao, J., Niu, M., Bai, H., Liang, X., Xu, H., Xu, C.: Pyramid r-cnn: Towards
  better performance and adaptability for 3d object detection.
\newblock In: ICCV (2021)

\bibitem{mao2021one}
Mao, J., Niu, M., Jiang, C., Liang, H., Chen, J., Liang, X., Li, Y., Ye, C.,
  Zhang, W., Li, Z., et~al.: One million scenes for autonomous driving: Once
  dataset.
\newblock arXiv preprint arXiv:2106.11037  (2021)

\bibitem{marsden2016fully}
Marsden, M., McGuinness, K., Little, S., O'Connor, N.E.: Fully convolutional
  crowd counting on highly congested scenes.
\newblock VISAPP  (2016)

\bibitem{marsden2017resnetcrowd}
Marsden, M., McGuinness, K., Little, S., O'Connor, N.E.: Resnetcrowd: A
  residual deep learning architecture for crowd counting, violent behaviour
  detection and crowd density level classification.
\newblock In: AVSS (2017)

\bibitem{meng2021count}
Meng, C., Liu, E., Neiswanger, W., Song, J., Burke, M., Lobell, D., Ermon, S.:
  Is-count: Large-scale object counting from satellite images with
  covariate-based importance sampling.
\newblock AAAI  (2021)

\bibitem{meng2021spatial}
Meng, Y., Zhang, H., Zhao, Y., Yang, X., Qian, X., Huang, X., Zheng, Y.:
  Spatial uncertainty-aware semi-supervised crowd counting.
\newblock In: ICCV (2021)

\bibitem{miao2020shallow}
Miao, Y., Lin, Z., Ding, G., Han, J.: Shallow feature based dense attention
  network for crowd counting.
\newblock In: AAAI (2020)

\bibitem{mo2020background}
Mo, H., Ren, W., Xiong, Y., Pan, X., Zhou, Z., Cao, X., Wu, W.: Background
  noise filtering and distribution dividing for crowd counting.
\newblock IEEE Transactions on Image Processing  (2020)

\bibitem{modolo2021understanding}
Modolo, D., Shuai, B., Varior, R.R., Tighe, J.: Understanding the impact of
  mistakes on background regions in crowd counting.
\newblock In: WACV (2021)

\bibitem{oh2020crowd}
Oh, M.h., Olsen, P.A., Ramamurthy, K.N.: Crowd counting with decomposed
  uncertainty.
\newblock In: AAAI (2020)

\bibitem{olmschenk2019dense}
Olmschenk, G., Chen, J., Tang, H., Zhu, Z.: Dense crowd counting convolutional
  neural networks with minimal data using semi-supervised dual-goal generative
  adversarial networks.
\newblock In: CVPR Workshops (2019)

\bibitem{olmschenk2018crowd}
Olmschenk, G., Tang, H., Zhu, Z.: Crowd counting with minimal data using
  generative adversarial networks for multiple target regression.
\newblock In: WACV (2018)

\bibitem{olmschenk2019improving}
Olmschenk, G., Tang, H., Zhu, Z.: Improving dense crowd counting convolutional
  neural networks using inverse k-nearest neighbor maps and multiscale
  upsampling.
\newblock VISAPP  (2019)

\bibitem{olmschenk2019generalizing}
Olmschenk, G., Zhu, Z., Tang, H.: Generalizing semi-supervised generative
  adversarial networks to regression using feature contrasting.
\newblock CVIU  (2019)

\bibitem{onoro2016towards}
Onoro-Rubio, D., L{\'o}pez-Sastre, R.J.: Towards perspective-free object
  counting with deep learning.
\newblock In: ECCV (2016)

\bibitem{pan2020attention}
Pan, X., Mo, H., Zhou, Z., Wu, W.: Attention guided region division for crowd
  counting.
\newblock In: ICASSP (2020)

\bibitem{pham2015count}
Pham, V.Q., Kozakaya, T., Yamaguchi, O., Okada, R.: Count forest: Co-voting
  uncertain number of targets using random forest for crowd density estimation.
\newblock In: ICCV (2015)

\bibitem{qiu2019crowd}
Qiu, Z., Liu, L., Li, G., Wang, Q., Xiao, N., Lin, L.: Crowd counting via
  multi-view scale aggregation networks.
\newblock In: ICME (2019)

\bibitem{rabaud2006counting}
Rabaud, V., Belongie, S.: Counting crowded moving objects.
\newblock In: CVPR (2006)

\bibitem{ranjan2018iterative}
Ranjan, V., Le, H., Hoai, M.: Iterative crowd counting.
\newblock In: ECCV (2018)

\bibitem{ranjan2020uncertainty}
Ranjan, V., Wang, B., Shah, M., Hoai, M.: Uncertainty estimation and sample
  selection for crowd counting.
\newblock In: ACCV (2020)

\bibitem{reddy2020few}
Reddy, M.K.K., Hossain, M., Rochan, M., Wang, Y.: Few-shot scene adaptive crowd
  counting using meta-learning.
\newblock In: WACV (2020)

\bibitem{reddy2021adacrowd}
Reddy, M.K.K., Rochan, M., Lu, Y., Wang, Y.: Adacrowd: unlabeled scene
  adaptation for crowd counting.
\newblock IEEE Transactions on Multimedia  (2021)

\bibitem{ren2020tracking}
Ren, W., Wang, X., Tian, J., Tang, Y., Chan, A.B.: Tracking-by-counting: Using
  network flows on crowd density maps for tracking multiple targets.
\newblock IEEE Transactions on Image Processing  (2020)

\bibitem{rong2021coarse}
Rong, L., Li, C.: Coarse-and fine-grained attention network with
  background-aware loss for crowd density map estimation.
\newblock In: WACV (2021)

\bibitem{sajid2021audio}
Sajid, U., Chen, X., Sajid, H., Kim, T., Wang, G.: Audio-visual transformer
  based crowd counting.
\newblock In: ICCV (2021)

\bibitem{sajid2021multi}
Sajid, U., Ma, W., Wang, G.: Multi-resolution fusion and multi-scale input
  priors based crowd counting.
\newblock In: ICPR (2021)

\bibitem{sajid2020plug}
Sajid, U., Wang, G.: Plug-and-play rescaling based crowd counting in static
  images.
\newblock In: WACV (2020)

\bibitem{sajid2022towards}
Sajid, U., Wang, G.: Towards more effective prm-based crowd counting via a
  multi-resolution fusion and attention network.
\newblock Neurocomputing  (2022)

\bibitem{sam2020completely}
Sam, D.B., Agarwalla, A., Joseph, J., Sindagi, V.A., Babu, R.V., Patel, V.M.:
  Completely self-supervised crowd counting via distribution matching.
\newblock arXiv preprint arXiv:2009.06420  (2020)

\bibitem{sam2020locate}
Sam, D.B., Peri, S.V., Sundararaman, M.N., Kamath, A., Babu, R.V.: Locate,
  size, and count: accurately resolving people in dense crowds via detection.
\newblock TPAMI  (2020)

\bibitem{sam2019almost}
Sam, D.B., Sajjan, N.N., Maurya, H., Babu, R.V.: Almost unsupervised learning
  for dense crowd counting.
\newblock In: AAAI (2019)

\bibitem{servadei2022label}
Servadei, L., Sun, H., Ott, J., Stephan, M., Hazra, S., Stadelmayer, T.,
  Lopera, D.S., Wille, R., Santra, A.: Label-aware ranked loss for robust
  people counting using automotive in-cabin radar.
\newblock In: ICASSP (2022)

\bibitem{shang2016end}
Shang, C., Ai, H., Bai, B.: End-to-end crowd counting via joint learning local
  and global count.
\newblock In: ICIP (2016)

\bibitem{shen2018crowd}
Shen, Z., Xu, Y., Ni, B., Wang, M., Hu, J., Yang, X.: Crowd counting via
  adversarial cross-scale consistency pursuit.
\newblock In: CVPR (2018)

\bibitem{shi2020real}
Shi, X., Li, X., Wu, C., Kong, S., Yang, J., He, L.: A real-time deep network
  for crowd counting.
\newblock In: ICASSP (2020)

\bibitem{shi2019counting}
Shi, Z., Mettes, P., Snoek, C.G.: Counting with focus for free.
\newblock In: ICCV (2019)

\bibitem{shi2018crowd}
Shi, Z., Zhang, L., Liu, Y., Cao, X., Ye, Y., Cheng, M.M., Zheng, G.: Crowd
  counting with deep negative correlation learning.
\newblock In: CVPR (2018)

\bibitem{shivapuja2021wisdom}
Shivapuja, S.V., Khamkar, M.P., Bajaj, D., Ramakrishnan, G., Sarvadevabhatla,
  R.K.: Wisdom of (binned) crowds: A bayesian stratification paradigm for crowd
  counting.
\newblock In: ACM Multimedia (2021)

\bibitem{sindagi2017cnn}
Sindagi, V.A., Patel, V.M.: Cnn-based cascaded multi-task learning of
  high-level prior and density estimation for crowd counting.
\newblock In: AVSS (2017)

\bibitem{sindagi2017generating}
Sindagi, V.A., Patel, V.M.: Generating high-quality crowd density maps using
  contextual pyramid cnns.
\newblock In: ICCV (2017)

\bibitem{sindagi2018survey}
Sindagi, V.A., Patel, V.M.: A survey of recent advances in cnn-based single
  image crowd counting and density estimation.
\newblock Pattern Recognition Letters  (2018)

\bibitem{sindagi2019ha}
Sindagi, V.A., Patel, V.M.: Ha-ccn: Hierarchical attention-based crowd counting
  network.
\newblock TIP  (2019)

\bibitem{sindagi2019multi}
Sindagi, V.A., Patel, V.M.: Multi-level bottom-top and top-bottom feature
  fusion for crowd counting.
\newblock In: ICCV (2019)

\bibitem{sindagi2020learning}
Sindagi, V.A., Yasarla, R., Babu, D.S., Babu, R.V., Patel, V.M.: Learning to
  count in the crowd from limited labeled data.
\newblock In: ECCV (2020)

\bibitem{sindagi2019pushing}
Sindagi, V.A., Yasarla, R., Patel, V.M.: Pushing the frontiers of unconstrained
  crowd counting: New dataset and benchmark method.
\newblock In: ICCV (2019)

\bibitem{song2021rethinking}
Song, Q., Wang, C., Jiang, Z., Wang, Y., Tai, Y., Wang, C., Li, J., Huang, F.,
  Wu, Y.: Rethinking counting and localization in crowds: A purely point-based
  framework.
\newblock In: ICCV (2021)

\bibitem{song2021choose}
Song, Q., Wang, C., Wang, Y., Tai, Y., Wang, C., Li, J., Wu, J., Ma, J.: To
  choose or to fuse? scale selection for crowd counting.
\newblock In: AAAI (2021)

\bibitem{tan2019crowd}
Tan, X., Tao, C., Ren, T., Tang, J., Wu, G.: Crowd counting via multi-layer
  regression.
\newblock In: ACM Multimedia (2019)

\bibitem{tang2022tafnet}
Tang, H., Wang, Y., Chau, L.P.: Tafnet: A three-stream adaptive fusion network
  for rgb-t crowd counting.
\newblock ISCAS  (2022)

\bibitem{teixeira2010survey}
Teixeira, T., Dublon, G., Savvides, A.: A survey of human-sensing: Methods for
  detecting presence, count, location, track, and identity.
\newblock ACM Computing Surveys  (2010)

\bibitem{thanasutives2021encoder}
Thanasutives, P., Fukui, K.i., Numao, M., Kijsirikul, B.: Encoder-decoder based
  convolutional neural networks with multi-scale-aware modules for crowd
  counting.
\newblock In: ICPR (2021)

\bibitem{walach2016learning}
Walach, E., Wolf, L.: Learning to count with cnn boosting.
\newblock In: ECCV (2016)

\bibitem{wan2019adaptive}
Wan, J., Chan, A.: Adaptive density map generation for crowd counting.
\newblock In: ICCV (2019)

\bibitem{wan2020modeling}
Wan, J., Chan, A.: Modeling noisy annotations for crowd counting.
\newblock NeurIPS  (2020)

\bibitem{wan2021fine}
Wan, J., Kumar, N.S., Chan, A.B.: Fine-grained crowd counting.
\newblock IEEE transactions on image processing  (2021)

\bibitem{wan2021generalized}
Wan, J., Liu, Z., Chan, A.B.: A generalized loss function for crowd counting
  and localization.
\newblock In: CVPR (2021)

\bibitem{wan2020kernel}
Wan, J., Wang, Q., Chan, A.B.: Kernel-based density map generation for dense
  object counting.
\newblock TPAMI  (2020)

\bibitem{wang2020distribution}
Wang, B., Liu, H., Samaras, D., Nguyen, M.H.: Distribution matching for crowd
  counting.
\newblock NeurIPS  (2020)

\bibitem{wang2021uniformity}
Wang, C., Song, Q., Zhang, B., Wang, Y., Tai, Y., Hu, X., Wang, C., Li, J., Ma,
  J., Wu, Y.: Uniformity in heterogeneity: Diving deep into count interval
  partition for crowd counting.
\newblock In: ICCV (2021)

\bibitem{wang2015deep}
Wang, C., Zhang, H., Yang, L., Liu, S., Cao, X.: Deep people counting in
  extremely dense crowds.
\newblock In: ACM Multimedia (2015)

\bibitem{wang2022hybrid}
Wang, F., Sang, J., Wu, Z., Liu, Q., Sang, N.: Hybrid attention network based
  on progressive embedding scale-context for crowd counting.
\newblock Information Sciences  (2022)

\bibitem{wang2019removing}
Wang, L., Yin, B., Tang, X., Li, Y.: Removing background interference for crowd
  counting via de-background detail convolutional network.
\newblock Neurocomputing  (2019)

\bibitem{wang2022stnet}
Wang, M., Cai, H., Han, X., Zhou, J., Gong, M.: Stnet: Scale tree network with
  multi-level auxiliator for crowd counting.
\newblock IEEE Transactions on Multimedia  (2022)

\bibitem{wang2021interlayer}
Wang, M., Cai, H., Zhou, J., Gong, M.: Interlayer and intralayer scale
  aggregation for scale-invariant crowd counting.
\newblock Neurocomputing  (2021)

\bibitem{wang2022crowdmlp}
Wang, M., Zhou, J., Cai, H., Gong, M.: Crowdmlp: Weakly-supervised crowd
  counting via multi-granularity mlp.
\newblock arXiv preprint arXiv:2203.08219  (2022)

\bibitem{wang2020mobilecount}
Wang, P., Gao, C., Wang, Y., Li, H., Gao, Y.: Mobilecount: An efficient
  encoder-decoder framework for real-time crowd counting.
\newblock Neurocomputing  (2020)

\bibitem{wang2022crowd}
Wang, Q., Breckon, T.P.: Crowd counting via segmentation guided attention
  networks and curriculum loss.
\newblock IEEE Transactions on Intelligent Transportation Systems  (2022)

\bibitem{wang2020nwpu}
Wang, Q., Gao, J., Lin, W., Li, X.: Nwpu-crowd: A large-scale benchmark for
  crowd counting and localization.
\newblock TPAMI  (2020)

\bibitem{wang2019learning}
Wang, Q., Gao, J., Lin, W., Yuan, Y.: Learning from synthetic data for crowd
  counting in the wild.
\newblock In: CVPR (2019)

\bibitem{wang2020pixel}
Wang, Q., Gao, J., Lin, W., Yuan, Y.: Pixel-wise crowd understanding via
  synthetic data.
\newblock IJCV  (2020)

\bibitem{wang2021pixel}
Wang, Q., Gao, J., Lin, W., Yuan, Y.: Pixel-wise crowd understanding via
  synthetic data.
\newblock IJCV  (2021)

\bibitem{wang2021neuron}
Wang, Q., Han, T., Gao, J., Yuan, Y.: Neuron linear transformation: Modeling
  the domain shift for crowd counting.
\newblock TNNLS  (2021)

\bibitem{wang2020density}
Wang, Q., Lin, W., Gao, J., Li, X.: Density-aware curriculum learning for crowd
  counting.
\newblock IEEE Transactions on Cybernetics  (2020)

\bibitem{wang2021self}
Wang, Y., Hou, J., Hou, X., Chau, L.P.: A self-training approach for
  point-supervised object detection and counting in crowds.
\newblock IEEE Transactions on Image Processing  (2021)

\bibitem{wang2021dense}
Wang, Y., Hou, X., Chau, L.P.: Dense point prediction: A simple baseline for
  crowd counting and localization.
\newblock In: ICMEW (2021)

\bibitem{wang2022eccnas}
Wang, Y., Ma, Z., Wei, X., Zheng, S., Wang, Y., Hong, X.: Eccnas: Efficient
  crowd counting neural architecture search.
\newblock TOMM  (2022)

\bibitem{wang2004image}
Wang, Z., Bovik, A.C., Sheikh, H.R., Simoncelli, E.P.: Image quality
  assessment: from error visibility to structural similarity.
\newblock TIP  (2004)

\bibitem{wang2018defense}
Wang, Z., Xiao, Z., Xie, K., Qiu, Q., Zhen, X., Cao, X.: In defense of
  single-column networks for crowd counting.
\newblock BMVC  (2018)

\bibitem{wei2020mspnet}
Wei, B., Yuan, Y., Wang, Q.: Mspnet: Multi-supervised parallel network for
  crowd counting.
\newblock In: ICASSP (2020)

\bibitem{wei2019boosting}
Wei, X., Du, J., Liang, M., Ye, L.: Boosting deep attribute learning via
  support vector regression for fast moving crowd counting.
\newblock Pattern Recognition Letters  (2019)

\bibitem{wen2021detection}
Wen, L., Du, D., Zhu, P., Hu, Q., Wang, Q., Bo, L., Lyu, S.: Detection,
  tracking, and counting meets drones in crowds: A benchmark.
\newblock In: CVPR (2021)

\bibitem{wu2021dynamic}
Wu, Q., Wan, J., Chan, A.B.: Dynamic momentum adaptation for zero-shot
  cross-domain crowd counting.
\newblock In: ACM Multimedia (2021)

\bibitem{wu2020fast}
Wu, X., Xu, B., Zheng, Y., Ye, H., Yang, J., He, L.: Fast video crowd counting
  with a temporal aware network.
\newblock Neurocomputing  (2020)

\bibitem{wu2020counting}
Wu, X., Zheng, Y., Ye, H., Hu, W., Ma, T., Yang, J., He, L.: Counting crowds
  with varying densities via adaptive scenario discovery framework.
\newblock Neurocomputing  (2020)

\bibitem{wu2019adaptive}
Wu, X., Zheng, Y., Ye, H., Hu, W., Yang, J., He, L.: Adaptive scenario
  discovery for crowd counting.
\newblock In: ICASSP (2019)

\bibitem{wu2021cranet}
Wu, Z., Sang, J., Shi, Y., Liu, Q., Sang, N., Liu, X.: Cranet: Cascade residual
  attention network for crowd counting.
\newblock In: ICME (2021)

\bibitem{xiong2017spatiotemporal}
Xiong, F., Shi, X., Yeung, D.Y.: Spatiotemporal modeling for crowd counting in
  videos.
\newblock In: ICCV (2017)

\bibitem{xiong2019open}
Xiong, H., Lu, H., Liu, C., Liu, L., Cao, Z., Shen, C.: From open set to closed
  set: Counting objects by spatial divide-and-conquer.
\newblock In: ICCV (2019)

\bibitem{xu2022autoscale}
Xu, C., Liang, D., Xu, Y., Bai, S., Zhan, W., Bai, X., Tomizuka, M.: Autoscale:
  Learning to scale for crowd counting.
\newblock IJCV  (2022)

\bibitem{xu2021dilated}
Xu, W., Liang, D., Zheng, Y., Xie, J., Ma, Z.: Dilated-scale-aware
  category-attention convnet for multi-class object counting.
\newblock IEEE Signal Processing Letters  (2021)

\bibitem{xu2021crowd}
Xu, Y., Zhong, Z., Lian, D., Li, J., Li, Z., Xu, X., Gao, S.: Crowd counting
  with partial annotations in an image.
\newblock In: ICCV (2021)

\bibitem{yan2021towards}
Yan, Z., Li, P., Wang, B., Ren, D., Zuo, W.: Towards learning multi-domain
  crowd counting.
\newblock IEEE Trans. Circuits Syst. Video Technol  (2021)

\bibitem{yan2019perspective}
Yan, Z., Yuan, Y., Zuo, W., Tan, X., Wang, Y., Wen, S., Ding, E.:
  Perspective-guided convolution networks for crowd counting.
\newblock In: ICCV (2019)

\bibitem{yan2021crowd}
Yan, Z., Zhang, R., Zhang, H., Zhang, Q., Zuo, W.: Crowd counting via
  perspective-guided fractional-dilation convolution.
\newblock IEEE Transactions on Multimedia  (2021)

\bibitem{yang2020counting}
Yang, B., Zhan, W., Wang, N., Liu, X., Lv, J.: Counting crowds using a
  scale-distribution-aware network and adaptive human-shaped kernel.
\newblock Neurocomputing  (2020)

\bibitem{yang2018multi}
Yang, J., Zhou, Y., Kung, S.Y.: Multi-scale generative adversarial networks for
  crowd counting.
\newblock In: ICPR (2018)

\bibitem{yang2021class}
Yang, S.D., Su, H.T., Hsu, W.H., Chen, W.C.: Class-agnostic few-shot object
  counting.
\newblock In: WACV (2021)

\bibitem{yang2020embedding}
Yang, Y., Li, G., Du, D., Huang, Q., Sebe, N.: Embedding perspective analysis
  into multi-column convolutional neural network for crowd counting.
\newblock IEEE Transactions on Image Processing  (2020)

\bibitem{yang2020reverse}
Yang, Y., Li, G., Wu, Z., Su, L., Huang, Q., Sebe, N.: Reverse perspective
  network for perspective-aware object counting.
\newblock In: CVPR (2020)

\bibitem{yang2020weakly}
Yang, Y., Li, G., Wu, Z., Su, L., Huang, Q., Sebe, N.: Weakly-supervised crowd
  counting learns from sorting rather than locations.
\newblock In: ECCV (2020)

\bibitem{zand2022multiscale}
Zand, M., Damirchi, H., Farley, A., Molahasani, M., Greenspan, M., Etemad, A.:
  Multiscale crowd counting and localization by multitask point supervision.
\newblock In: ICASSP (2022)

\bibitem{zhang2019relational}
Zhang, A., Shen, J., Xiao, Z., Zhu, F., Zhen, X., Cao, X., Shao, L.: Relational
  attention network for crowd counting.
\newblock In: ICCV (2019)

\bibitem{zhang2019attentional}
Zhang, A., Yue, L., Shen, J., Zhu, F., Zhen, X., Cao, X., Shao, L.: Attentional
  neural fields for crowd counting.
\newblock In: ICCV (2019)

\bibitem{zhang2015cross}
Zhang, C., Li, H., Wang, X., Yang, X.: Cross-scene crowd counting via deep
  convolutional neural networks.
\newblock In: CVPR (2015)

\bibitem{zhang2018crowd}
Zhang, L., Shi, M., Chen, Q.: Crowd counting via scale-adaptive convolutional
  neural network.
\newblock In: WACV (2018)

\bibitem{zhang2019nonlinear}
Zhang, L., Shi, Z., Cheng, M.M., Liu, Y., Bian, J.W., Zhou, J.T., Zheng, G.,
  Zeng, Z.: Nonlinear regression via deep negative correlation learning.
\newblock TPAMI  (2019)

\bibitem{zhang2019wide}
Zhang, Q., Chan, A.B.: Wide-area crowd counting via ground-plane density maps
  and multi-view fusion cnns.
\newblock In: CVPR (2019)

\bibitem{zhang20203d}
Zhang, Q., Chan, A.B.: 3d crowd counting via multi-view fusion with 3d gaussian
  kernels.
\newblock In: AAAI (2020)

\bibitem{zhang2021cross}
Zhang, Q., Lin, W., Chan, A.B.: Cross-view cross-scene multi-view crowd
  counting.
\newblock In: CVPR (2021)

\bibitem{zhang2019multi}
Zhang, Y., Zhou, C., Chang, F., Kot, A.C.: Multi-resolution attention
  convolutional neural network for crowd counting.
\newblock Neurocomputing  (2019)

\bibitem{zhang2016single}
Zhang, Y., Zhou, D., Chen, S., Gao, S., Ma, Y.: Single-image crowd counting via
  multi-column convolutional neural network.
\newblock In: CVPR (2016)

\bibitem{zhao2019scale}
Zhao, M., Zhang, C., Zhang, J., Porikli, F., Ni, B., Zhang, W.: Scale-aware
  crowd counting via depth-embedded convolutional neural networks.
\newblock TCSVT  (2019)

\bibitem{zhao2020flow}
Zhao, Z., Han, T., Gao, J., Wang, Q., Li, X.: A flow base bi-path network for
  cross-scene video crowd understanding in aerial view.
\newblock In: ECCV (2020)

\bibitem{zhao2020active}
Zhao, Z., Shi, M., Zhao, X., Li, L.: Active crowd counting with limited
  supervision.
\newblock In: ECCV (2020)

\bibitem{zheng2021learning}
Zheng, L., Li, Y., Mu, Y.: Learning factorized cross-view fusion for multi-view
  crowd counting.
\newblock In: ICME (2021)

\bibitem{zhong2022improved}
Zhong, X., Yan, Z., Qin, J., Zuo, W., Lu, W.: An improved normed-deformable
  convolution for crowd counting.
\newblock SPL  (2022)

\bibitem{zhou2021locality}
Zhou, J.T., Zhang, L., Du, J., Peng, X., Fang, Z., Xiao, Z., Zhu, H.:
  Locality-aware crowd counting.
\newblock TPAMI  (2021)

\bibitem{zhou2018crowd}
Zhou, Q., Zhang, J., Che, L., Shan, H., Wang, J.Z.: Crowd counting with limited
  labeling through submodular frame selection.
\newblock IEEE Transactions on Intelligent Transportation Systems  (2018)

\bibitem{zhou2020adversarial}
Zhou, Y., Yang, J., Li, H., Cao, T., Kung, S.Y.: Adversarial learning for
  multiscale crowd counting under complex scenes.
\newblock IEEE transactions on cybernetics  (2020)

\bibitem{zhu2021graph}
Zhu, P., Peng, T., Du, D., Yu, H., Zhang, L., Hu, Q.: Graph regularized flow
  attention network for video animal counting from drones.
\newblock IEEE Transactions on Image Processing  (2021)

\bibitem{zhu2018visdrone}
Zhu, P., Wen, L., Du, D., Bian, X., Ling, H., Hu, Q., Wu, H., Nie, Q., Cheng,
  H., Liu, C., et~al.: Visdrone-vdt2018: The vision meets drone video detection
  and tracking challenge results.
\newblock In: ECCV Workshops (2018)

\bibitem{zitouni2016advances}
Zitouni, M.S., Bhaskar, H., Dias, J., Al-Mualla, M.E.: Advances and trends in
  visual crowd analysis: A systematic survey and evaluation of crowd modelling
  techniques.
\newblock Neurocomputing  (2016)

\bibitem{zou2019attend}
Zou, Z., Cheng, Y., Qu, X., Ji, S., Guo, X., Zhou, P.: Attend to count: Crowd
  counting with adaptive capacity multi-scale cnns.
\newblock Neurocomputing  (2019)

\bibitem{zou2020crowd}
Zou, Z., Liu, Y., Xu, S., Wei, W., Wen, S., Zhou, P.: Crowd counting via
  hierarchical scale recalibration network.
\newblock ECAI  (2020)

\bibitem{zou2019enhanced}
Zou, Z., Shao, H., Qu, X., Wei, W., Zhou, P.: Enhanced 3d convolutional
  networks for crowd counting.
\newblock BMVC  (2019)

\end{thebibliography}

\end{document}